\documentclass[
]{ceurart}

\usepackage{listings}
\newcommand{\norm}[1]{\left\lVert#1\right\rVert}

\usepackage[textsize=tiny]{todonotes}

\begin{document}

\copyrightyear{2026}
\copyrightclause{Copyright for this paper by its authors.
  Use permitted under Creative Commons License Attribution 4.0
  International (CC BY 4.0).}

\conference{CMNA'26: 26th Workshop on Computational Models of Natural Argument, September 14--15, 2026, Barcelona}

\title{Embedding Models for Stance-Aware Argument Retrieval}

\author[1]{Angelo Sparacino}[%
email=angelojsparacino@outlook.com,
orcid=0009-0005-0126-7836
]
\author[1]{Francesca Toni}[%
email=ft@imperial.ac.uk,
orcid=0000-0001-8194-1459
]
\author[1]{Adam Dejl}[%
email=adam.dejl18@imperial.ac.uk,
orcid=0009-0006-0274-4160
]

\address[1]{Department of Computing, Imperial College London, UK}


\begin{abstract}
  In computational argumentation, obtaining arguments that explicitly support or attack given claims is a critical precursor to downstream reasoning tasks. When these supporting and attacking arguments are to be retrieved using semantic search methods, they need to be assessed for topic-relevance to the claims of interest as well as for correctness of their (positive or negative) stance towards the claims. In this paper we explore how dense \emph{embedding models} (hereafter, models), powering modern retrieval pipelines, can serve as the basis of semantic search incorporating this dual assessment. We show experimentally that existing models struggle with asymmetric reasoning, exhibiting a strong bias toward topical overlap while ignoring instructional stance. We also show that correcting this bias via contrastive training triggers a new failure mode where models over-correct, over-fixating on polarity keywords (e.g., ``supports'' or ``refutes'') at the expense of the semantic topic. We thus introduce diagnostic word-ablation metrics to quantify this phenomenon and propose a data-centric solution. By implementing a balanced argument curriculum alongside LLM-augmented, stance-inverted arguments, we force the (embedding) models to learn deeper directional logic rather than exploiting superficial lexical shortcuts. Our evaluation demonstrates that, for sufficiently powerful models, this approach can alleviate the observed overcorrection, achieving further improvements in stance-aware argument retrieval.
\end{abstract}


\maketitle

\section{Introduction}

The modern digital landscape is associated with an unprecedented proliferation of data. To process this information, the field of computational argumentation relies heavily on automated argument mining pipelines \cite{lawrence-reed-2019-argument}. A foundational prerequisite to these tasks is the accurate retrieval of evidence that explicitly supports or attacks a claim. While Large Language Models (LLMs) are often employed in Retrieval-Augmented Generation (RAG) \cite{3495724.3496517} to ground their reasoning in external evidence, the efficacy of these pipelines is bottlenecked by their retrieval component.

\begin{figure}
    \centering
    \includegraphics[width=0.5\linewidth]{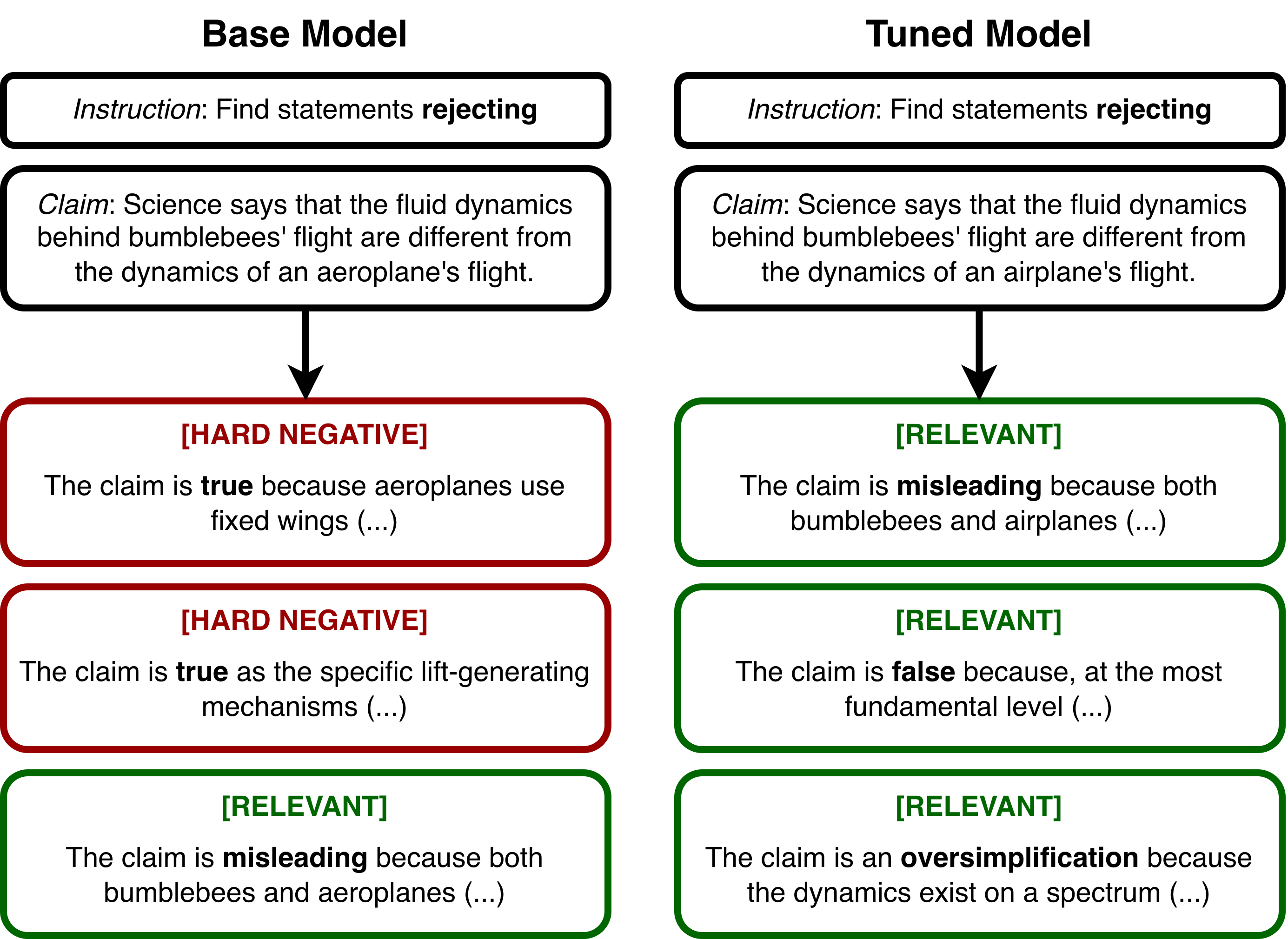}
    \caption{An illustration of the stance-aware argument retrieval problem addressed in this work. The Base Model (left) successfully matches on the semantic topic for all retrieved arguments but ignores the stance instruction (`rejecting'), resulting in the retrieval of two `hard negatives'. Our Tuned Model (right) successfully overcomes this bias, retrieving topically relevant arguments with the correct opposing stance.}
    \label{fig:intro_fig}
\end{figure}

Modern dense retrievers utilise a bi-encoder architecture to map textual sequences into a continuous, high-dimensional vector space \cite{karpukhin-etal-2020-dense}, quantifying semantic relevance through spatial proximity. However, applying pre-trained embedding models (hereafter, models) to the asymmetric demands of (attacking/supporting) argument retrieval demonstrates that they struggle on this task. Indeed, while instruction-tuned embedding models \cite{su-etal-2023-one, qwen3embedding, bge_embedding} excel at fetching documents based on general subject matter, their internal mechanisms often default to superficial topical matching. These models heavily rely on semantic overlap while largely ignoring relational stance \cite{sinha-etal-2021-unnatural}, as illustrated in Figure \ref{fig:intro_fig} (left).
Here, given an instruction such as ``Find statements rejecting'' and a claim relating to some topic, models frequently retrieve arguments that perfectly match the topic but directly violate the instruction. 

Having evaluated this problem experimentally (Section~\ref{sec:baseline_model_evaluation}), we initially
 attempt to correct this behaviour through contrastive tuning, but observe that this may cause the embedding space to undergo a phenomenon we call `topical collapse'. During optimisation, the model shifts its representational capacity away from the underlying subject matter and over-fixates on explicit polarity keywords.

To address this problem as well as the general limitations of embedding models on the stance-aware argument retrieval task, we investigate several targeted interventions aiming to achieve balance between topic- and stance-sensitivity (Section~\ref{sec:solution}). The primary contributions of this work are as follows:
\begin{itemize}
    \item We conduct an extensive experimental evaluation of embedding models' performance on the stance-aware argument retrieval task, including through novel diagnostic metrics of Relative Instruction/Claim Sensitivity (RIS/RCS) and Directional Impact (DI). The metrics quantify the effects of specific input text portions on the resulting similarity through word ablation, providing deeper insight about model focus and the topical collapse phenomenon.
    \item We introduce targeted interventions for stance-aware retrieval, aiming to improve the performance of embedding models on this task while mitigating topical collapse. This includes devising a refined fine-tuning curriculum as well as exploring hybrid search approaches.
\end{itemize}

Overall, our experiments show that while base embedding models perform poorly on stance-aware retrieval, their performance can be substantially improved through targeted fine-tuning. To enable reuse by the research community, our codebase, datasets and models are publicly available\footnote{Code: \url{https://github.com/as9122/cmna26-stance-aware-text-retrieval}, datasets and models: \url{https://huggingface.co/collections/as9122/cmna26-stance-aware-text-retrieval}.}.

\section{Related Work}

\noindent\textbf{Argument Mining}\quad
Argument mining~\cite{DBLP:journals/coling/LawrenceR19} aims to identify, in text, arguments, their components, and dialectical relations (notably of attack or support) among them. In this paper we focus on relation-based argument mining~\cite{DBLP:conf/naacl/CarstensT15}, specifically on retrieving arguments supporting or attacking given claims. Rather than using LLMs to perform this task, e.g., as in \cite{DBLP:conf/coling/Gorur0T25}, we rely on embedding models.

\noindent\textbf{Dense Retrieval and Foundation Models}\quad
Modern retrieval systems rely on the bi-encoder architecture to semantically match queries to related documents. In this framework, a query and a document are processed independently by neural embedding models to generate high-dimensional, real-valued dense vectors. This allows relevance to be evaluated mathematically, typically via cosine similarity, and enables computationally efficient offline indexing \cite{xu-2026-survey-ir}.

However, static document embeddings introduce a ``representational bottleneck'' \cite{nogueira2020passagererankingbert} because they must compress all possible semantic interpretations into a single vector. To bypass this limitation, instruction-tuned models like Instructor \cite{su-etal-2023-one}, and more recently, decoder-only foundation models \cite{qwen3embedding}, were developed. By prepending a natural language instruction to the query, these architectures condition the embedding space to prioritise specific semantic features based on explicit user intent.

To train these foundation models for retrieval tasks, the standard industry practice relies on contrastive learning methodologies such as Multiple Negatives Ranking Loss (MNRL) \cite{henderson2017efficientnaturallanguageresponse}, an adaptation of InfoNCE \cite{oord2019representationlearningcontrastivepredictive}. During training, MNRL forces the model to maximise similarity between a query and a positive document, while simultaneously minimising similarity with a batch of negative documents.

\noindent\textbf{Lexical Bias and Instruction-Aware Retrieval}\quad
While the architectural transition to foundation models suggests an inherent capacity for complex reasoning, empirical stress testing reveals that models often regress to a ``bag-of-words'' heuristic. Rather than processing an instruction as a logical constraint, models treat it merely as a source of additional keywords to be matched \cite{DBLP:journals/corr/abs-2402-14334}. Benchmarks such as FollowIR \cite{weller-etal-2025-followir} and InstructIR \cite{DBLP:journals/corr/abs-2402-14334} have quantified this weakness, demonstrating the sensitivity of retrieval models to phrasing and surface-level lexical patterns.

In the context of text retrieval, subsets of documents can also act as ``hard negatives'' \cite{DBLP:journals/corr/abs-2505-21439}. For instruction-tuned models, these are often documents that share high topical overlap with the query but explicitly violate the instruction (e.g., a document refuting climate change when instructed to find supporting arguments). Because the embedding space is dominated by lexical overlap, the ``negative'' document often contains an equal or higher density of query terms than a true ``positive'' document, causing its embedding to be undesirably close to the query. A common approach to address this issue is hard-negative mining, where a retrieval model is intentionally exposed to hard-negative documents during training \cite{karpukhin-etal-2020-dense, DBLP:conf/iclr/XiongXLTLBAO21,moreira-2025-hard-negative}. This strategy can be optionally combined with other data-centric methods such as balanced sampling \cite{10.1145/3404835.3462891}
 and LLM data augmentation \cite{DBLP:journals/corr/abs-2202-05144, DBLP:conf/iclr/DaiZMLNLBGHC23, DBLP:conf/nips/Ouyang0JAWMZASR22, DBLP:conf/nips/ZhengC00WZL0LXZ23}.
 
\section{Task Formulation and Experimental Setup}

In this section, we provide an overview of the considered task of stance-aware argument retrieval while also describing the general experimental setup and evaluation metrics.  This background will be helpful for our later analysis evaluating the performance of state-of-the-art embedding models on stance-aware argument retrieval as well as our investigation of techniques to improve this performance.

\subsection{Task Definition}
We consider the task of stance-aware argument retrieval, a search problem where an embedding model is used to retrieve relevant documents (arguments) from a background corpus, taking into account both the semantic topic of a target claim and, crucially, directional stance. Such stance can amount to either \emph{supporting} or \emph{attacking} the claim, in line with the two relation types in bipolar argumentation frameworks \cite{cayrol-2005-bipolar}. In particular, we define an input query $Q$ as the concatenation of an instruction $I$, specifying the stance constraint (e.g., ``Find arguments supporting...'') and a claim $C$, specifying the core subject matter, such that $Q = I \oplus C$. This query can then be encoded using an instruction-tuned embedding model $\mathcal{M}$ and compared against the embeddings of documents in the background corpus $\mathcal{D}$, with those most similar in terms of cosine similarity $\text{sim}_{\mathcal{M}}$\footnote{In the subsequent text, we will drop the lower-index $\mathcal{M}$ when not referring to a specific embedding model.} being retrieved. Here, $\text{sim}_{\mathcal{M}}$ is defined as:
\begin{equation*}
    \text{sim}_{\mathcal{M}}(Q, D) = \frac{\mathcal{M}(Q) \cdot \mathcal{M}(D)}{\norm{\mathcal{M}(Q)}{\norm{\mathcal{M}(D)}}}
\end{equation*}

Given the requirements of the task, any document $D \in \mathcal{D}$ falls into one of four categories:
\begin{itemize}
    \item Positive $(D_\text{pos})$: Matches both the claim topic and the requested stance.
    \item Hard Negative $(D_\text{hn})$: Matches the claim topic but directly violates the requested stance.
    \item Semi-Hard Negative $(D_\text{shn})$: Does not match the claim topic but matches the requested stance.
    \item Easy Negative $(D_\text{en})$: Does not match the claim topic nor the requested stance.
\end{itemize}
In our experiments, we treat arguments associated with the same claim but opposite stance as hard negatives, arguments associated with different claims but matching the requested stance as semi-hard negatives, and arguments associated with different claims and opposite stance as easy negatives.

The primary challenge of stance-aware argument retrieval lies in distinguishing between $D_\text{pos}$ and $D_\text{hn}$, as both share a high degree of lexical overlap with the underlying claim $C$.

\subsection{Datasets and Setup}
All our experiments maintain a strict separation between training, validation and evaluation resources. For training, we utilised an 80\% split of the TFU Training Arguments dataset \cite{DBLP:journals/corr/abs-2605-20098}, paired with 20 standardised instructions (10 for supporting and 10 for attacking arguments) to ensure generalisation across varied phrasing. The remaining 20\% held-out split, alongside two TFU Validation Arguments datasets, generated by GPT-5 and Qwen3-8B, respectively \cite{DBLP:journals/corr/abs-2605-20098}, were used for initial in-domain evaluation. Finally, out-of-domain capabilities were evaluated using the TFU Evaluation Arguments dataset \cite{DBLP:journals/corr/abs-2605-20098}, the AVeriTeC Arguments dataset \cite{DBLP:conf/nips/SchlichtkrullG023, DBLP:journals/corr/abs-2605-20098}, and ArgTumour, a small domain-specific medical dataset focused on glioblastoma treatments \cite{10.1093/neuonc/noaf185.104}. To better assess generalisation, we also used additional unseen instructions (5 supporting and 5 attacking) for validation and evaluation. All instructions are provided in Appendix \ref{app:instr_lexicon}. Dataset statistics and further details on our training setup are given in Appendix \ref{app:experimental_details}.

To power experiments involving contrastive tuning, a triplet generation engine produced training triplets consisting of a query, a positive document and a negative document. A custom batch sampler was also utilised to ensure all triplets within a single training batch featured distinct claims, eliminating the risk of false in-batch negatives. In all our fine-tuning experiments, we used Low-Rank Adaptation (LoRA) \cite{hu-2022-lora} with a rank of $r = 64$, reducing the computational costs and enabling us to perform a wider range of experiments. To assess the generalisation of the proposed methods, we considered three pre-trained embedding models: BGE-Large \cite{bge_embedding}, Instructor-XL \cite{su-etal-2023-one} and Qwen3-Embedding-8B \cite{qwen3embedding}.

\subsection{Evaluation Metrics}
We evaluate retrieval performance using \textbf{Precision@R} as our primary metric, which dynamically scales the maximum achievable score based on the total number of positive documents $R$ available for a query, as well as \textbf{NDCG@10}, which applies a logarithmic discount to penalise poorly ranked results.

To explain the behaviour of the considered embedding models and diagnose the underlying cause of the retrieval failures, we employ a structured word ablation protocol. Let $\Delta_w$ denote the raw change in cosine similarity between the query $Q$ and document $D$ when a single word $w \in Q$\footnote{We assume that $w$ carries its index so that $Q$, $I$, $C$ and $D$ are ordered and repeated words are distinct.} is ablated:
\begin{equation*}
    \Delta_w(Q,D) = \text{sim}(Q \setminus \{w\},D) - \text{sim}(Q,D)
\end{equation*}
Building upon this, we define two diagnostic metrics:
\begin{itemize}
    \item \textbf{Relative Instruction/Claim Sensitivity (RIS/RCS)}: This measures the relative sensitivity of the model-driven similarity measure to the instruction and claim components of the query:
    \begin{equation*}
        \text{RIS}(Q,D) = \frac{\sum_{w\in I}\left|\Delta_{w}(Q,D)\right|}{\sum_{w\in Q}\left|\Delta_{w}(Q,D)\right|+\epsilon} , \quad \text{RCS}(Q,D) = \frac{\sum_{w\in C}|\Delta_{w}(Q,D)|}{\sum_{w\in Q}|\Delta_{w}(Q,D)|+\epsilon}
    \end{equation*}
    High RIS values indicate that the model is predominantly focusing on the instruction specifying the desired argumentative stance, while high RCS values indicate that the model is mainly focused on the claim topic, with $RIS(Q, D) + RCS(Q, D) \approx 1$.
    A small constant $\epsilon$ prevents division by zero.
    \item \textbf{Directional Impact (DI)}: This averages the raw, signed $\Delta_w$ values strictly for the subset of stance keywords $\mathcal{S}_I$ (e.g., ``supporting'', ``attacking'', etc) to test the model's relational logic:
    \begin{equation*}
        \text{DI}(Q,D) = \frac{1}{|\mathcal{S}_{I}|} \sum_{w\in\mathcal{S}_{I}} \Delta_{w}(Q,D)
    \end{equation*}
    For a hard negative $D_\text{hn}$, removing the stance word causes the instruction to become stance-agnostic, which should typically result in a higher similarity score and $\text{DI} > 0$.
\end{itemize}

We note that these metrics are solely based on the input-output behaviour of the model during word ablation rather than its internal representations. While this makes the metrics more widely applicable, it also causes them to be potentially sensitive to syntactic changes after ablations.

\section{Baseline Model Evaluation}
\label{sec:baseline_model_evaluation}

\subsection{Baseline Performance}

Given the considered stance-aware retrieval task, we first evaluate the performance of base instruction-following embedding models and their counterparts optimised using a baseline fine-tuning strategy. This strategy optimises the models using a homogeneous curriculum comprised of hard-negative triplets using MNRL. The results (in Figure \ref{fig:base_retrieval_metrics}) show substantial performance differences between the models:

\begin{figure}
    \centering
    \includegraphics[width=0.9\linewidth]{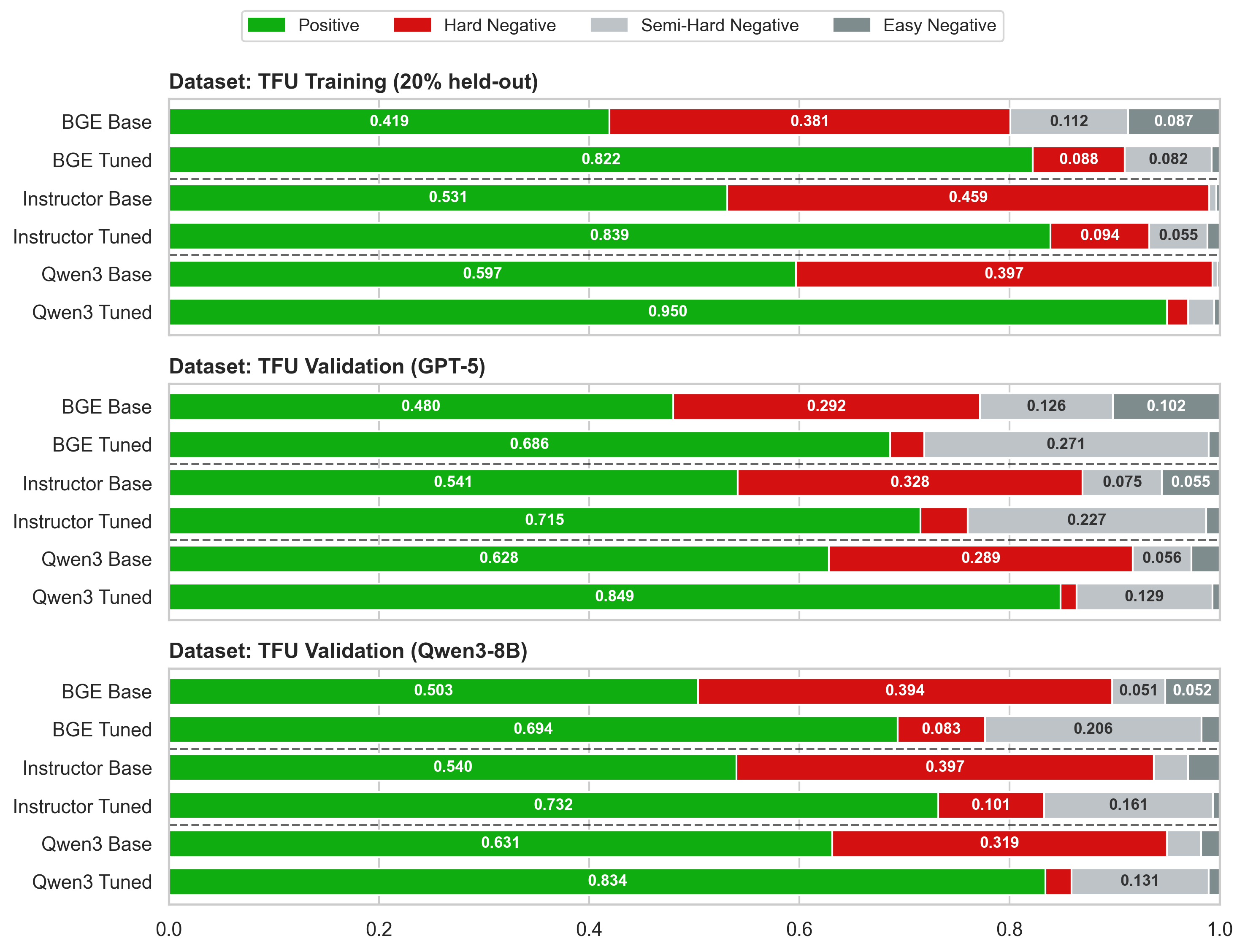}
    \caption{Average Precision@R for base and tuned models evaluated on unseen instructions. The base models exhibit severe instruction blindness, retrieving a high proportion of stance-violating hard negatives. While fine-tuning successfully reduces hard-negative retrievals, it also increases the retrieval of semi-hard negatives. This suggests that the used contrastive curriculum forces the models to overly focus on instructional stance at the expense of the claim topic, resulting in a phenomenon that we call topical collapse.}
    \label{fig:base_retrieval_metrics}
\end{figure}

\begin{itemize}

\item \textbf{BGE-Large}: Fine-tuning substantially improved BGE-Large's in-domain Precision@R on the held-out TFU Training data from 0.419 to 0.822, reducing stance error to 8.8\%. However, it remained the lowest-performing architecture overall, exhibiting a pronounced spike in semi-hard negatives on the TFU Validation sets (constituting 27.1\% of the retrieved items on GPT-5 data).

\item \textbf{Instructor-XL}: Instructor-XL benefited from fine-tuning, increasing its Precision@R on the held-out TFU Training data partition from 0.531 to 0.839 and dropping its stance error\footnote{We use the term \emph{stance error} interchangeably with hard negative retrieval rate or Hard@R, since hard negatives match the claim topic but violate the requested stance.} to 9.4\%.

\item \textbf{Qwen3-Embedding-8B}: Qwen3-Embedding-8B demonstrated the overall best performance. On the held-out training data partition, the fine-tuned model achieved a Precision@R of 0.95, reducing the originally high stance error down from 39.7\% to 2\%.

\end{itemize}

Overall, we find that none of the evaluated base embedding models perform stance-aware argument retrieval reliably, despite being instructed about the desired argument stance. While fine-tuning successfully suppressed stance errors for Qwen and Instructor-XL, it also triggered an associated increase in retrieving semi-hard negatives (documents matching the stance but not the topic). On the GPT-5 validation split, Qwen's semi-hard negative rate more than doubled, from 5.6\% to 12.9\%. For Instructor-XL, this failure was even more pronounced, increasing from 7.5\% to 22.6\%. These results suggest that standard contrastive optimisation against a homogeneous dataset (i.e., with all triplet negatives being hard negatives) forces models to over-correct, unduly focusing on argument stance while partially neglecting the topic. We elaborate on this phenomenon in the following subsection.

In an additional experiment, we test whether the general shortcomings of bi-encoder embedding models can be mitigated by using an off-the-shelf cross-encoder reranker. Cross-encoders compute semantic similarity scores by applying full self-attention across the concatenated query and document, and can be used to refine the ranking of documents originally retrieved by a bi-encoder. Such a setup is commonly used to improve the reliability of retrieval pipelines \cite{pandit-2026-rerankers}. However, empirical evaluation demonstrates that the off-the-shelf reranker fails at our task of stance-aware argument retrieval. When applied to the base Qwen3-Embedding-8B base model, the reranker, Qwen3-Reranker-8B, yielded negligible improvements, with stance errors remaining high (dropping marginally from 39.7\% to 38.5\%). More alarmingly, passing a candidate pool from the fine-tuned bi-encoder model to the base reranker actively counteracted the post-fine-tuning performance improvements. Because the cross-encoder relies on its own pre-trained lexical biases, its application reduced Precision@R on TFU Training from 0.95 to 0.645 and increased the stance error to 35.3\%. These results show that even the application of a computationally demanding reranker may be insufficient for achieving high performance on stance-aware argument retrieval, prompting us to focus on improving bi-encoder models directly.

Full experimental results, including exact NDCG@10 scores and standard deviations across all models and datasets, are provided in Appendix \ref{app:retrieval_results}.

\subsection{Topical Collapse}
\label{sec:topical-collapse}
While the retrieval metrics demonstrated a clear reduction in stance error for models like Qwen and Instructor-XL after fine-tuning, the associated rise in semi-hard negatives warrants further investigation. To diagnose the cause, we use our word ablation metrics to evaluate word-level similarity shifts. Comprehensive results for word ablation metrics across all models/datasets are given in Appendix \ref{app:ablation_metrics_tables}.

First, we evaluate the Directional Impact (DI) to confirm that training succeeded in making models more sensitive to stance logic. In the base models, ablating instruction stance verbs caused a near-zero similarity shift on average (e.g., 0.036 for hard negatives from TFU Training for Qwen), showing the base models' tendency to largely ignore relational constraints. Conversely, after fine-tuning, the average DI for hard negatives has substantially increased (e.g., to 0.238 on TFU Training for Qwen) while the average DI for positive arguments has substantially decreased (e.g., to -0.199 on TFU Training for Qwen). This indicates that, for the fine-tuned models, the stance keywords cause the query embeddings to draw substantially closer to the positive arguments while moving further away from the hard negatives.

However, evaluating the model's RIS and RCS reveals the drawbacks of this stance focus. Considering Qwen as a representative example, the base version of the model exhibited a strong claim bias, resulting in an average RCS of 75.2\% on the TFU Training Arguments dataset. Following contrastive fine-tuning, the model's average RIS on the dataset increased to 36\%, with the corresponding RCS decrease to 64\%. This suggests that fine-tuning has shifted model's focus to the stance instruction, but given the associated increase in semi-hard negative retrieval, this likely came at the cost of accurately representing the topic. This is corroborated by the full RCS distribution visualised in Figure \ref{fig:train_rcs}, which shows a substantially heavier lower tail for the fine-tuned model (see Appendix \ref{app:ablation_metrics_figs} for more related figures). Individual word-ablation heatmaps, such as the example in Appendix \ref{app:ablation_heatmap}, provide additional visualisation of the stance-topic trade-off, demonstrating how models shift focus from topical nouns to stance verbs.

These findings suggest that undue focus on stance may be detrimental to assessing topical relevance. The model successfully learns to retrieve attacking arguments but loses some precision in representing the topic of the claim being attacked. We refer to this failure mode as the \emph{topical collapse}.

\section{Targeted Interventions for Stance-Aware Retrieval}
\label{sec:solution}
\subsection{Rationale}
The analysis in Section \ref{sec:topical-collapse} established that while contrastive fine-tuning successfully forces the bi-encoder to account for instruction constraints, it inadvertently triggers topical collapse. We hypothesise this is due to the used data mixture, with the embedding space shaped by the optimisation pressure to differentiate between the positive and hard negative samples. This pressure may result in interference \cite{mccloskey-1989-catastrophic-forgetting}, reducing the ability of the model to accurately represent the semantic topic as a by-product of becoming more sensitive to the stance. Additionally, the training may be negatively affected by spurious patterns in the training set, as positive and hard-negative documents commonly share similar vocabulary and thematic subject matter. Therefore, we posit that increasing the semantic diversity of the training data and reducing these spurious differences may mitigate topical collapse.

\subsection{Data-Centric Interventions}

\noindent\textbf{Mixed Dataset}\quad The triplet negatives in the baseline, homogeneous curriculum consisted solely of hard negatives $(D_\text{hn})$. To test whether this strictly homogeneous training data caused topical collapse, we introduce a `mixed' curriculum. In this distribution, the negative sample space is partitioned equally with one-third hard negatives $(D_\text{hn})$, one-third semi-hard negatives $(D_\text{shn})$ and one-third easy negatives $(D_\text{en})$. To isolate the effect of the data distribution, the number of triplets was fixed at the baseline ceiling of 31,800. Reintroducing $D_\text{shn}$ forces the model to preserve its topical sensitivity.

\noindent\textbf{Stance-Inverted Argument Augmentation}\quad While the mixed curriculum mitigates topical collapse, the remaining $D_\text{hn}$ triplets still possess different sentence structures and lexical patterns compared to the positive documents ($D_\text{pos}$). To prevent the optimiser from exploiting these spurious differences, we generate synthetic, stance-inverted hard negatives to ensure near-identical sentence structure and substantial lexical overlap. However, naively prompting an LLM to invert an argument introduces new risks. For example, simply negating the arguments may make them inconsistent with real-world knowledge. Pre-trained embedding models, particularly those based on generative backbones, may identify such inconsistencies and use them as a spurious signal.

To construct helpful synthetic hard negatives, our generation pipeline uses a `concede and sever' strategy. The generating LLM concedes any factual knowledge, but synthetically severs the causal link to the target claim while minimising lexical differences. To operationalise this, Gemini 3.5 Flash was instructed to generate candidate stance inversions using a one-shot prompt. The complete prompt is provided in Appendix \ref{app:stance_inversion_prompt}. We then replaced 50\% of the natural hard negatives in the mixed curriculum with their synthetic counterparts. The $D_\text{shn}$ and $D_\text{en}$ data portions were left unchanged.

To validate this approach, we measured word-level Jaccard similarity. Standard argument pairs exhibited an average overlap of just 0.0938. In contrast, our synthetic pairs achieved 0.5866 (0.5896 with NLTK lemmatisation). While this falls short of the 90\% target specified in the generation prompt, the discrepancy is likely an artefact of the Jaccard metric (intersection over union). Adding even a few words to invert the stance heavily penalises the score. Nonetheless, this six-fold increase confirms the successful generation of lexically similar arguments.

\subsection{Hybrid Search}
While the data-centric interventions attempt to improve the dense encoder's ability to balance topic and stance, vector embeddings inherently struggle with lexical matching for rare, domain-specific entities, such as complex medical terms in the ArgTumour dataset. To address this, we implemented a hybrid search pipeline based on a `division of responsibility'. We fuse our tuned dense retriever with a traditional sparse lexical retriever (BM25 utilising local TF-IDF). Because the sparse model excels at lexical anchoring but is entirely stance-blind, and the tuned dense model excels at relational stance logic, combining them aims to result in more robust retrieval. We utilised Relative Score Fusion (RSF) with a conservative sparse weighting $(\lambda \in [0.1, 0.15, 0.2])$. This configuration allows the BM25 component to act as a lightweight topical filter while heavily relying on the tuned bi-encoder for stance constraints.

\section{Results}

\subsection{Intervention Evaluation}
\label{subsec:intervention_evaluation}
\begin{figure}
    \centering
    \includegraphics[width=0.9\linewidth]{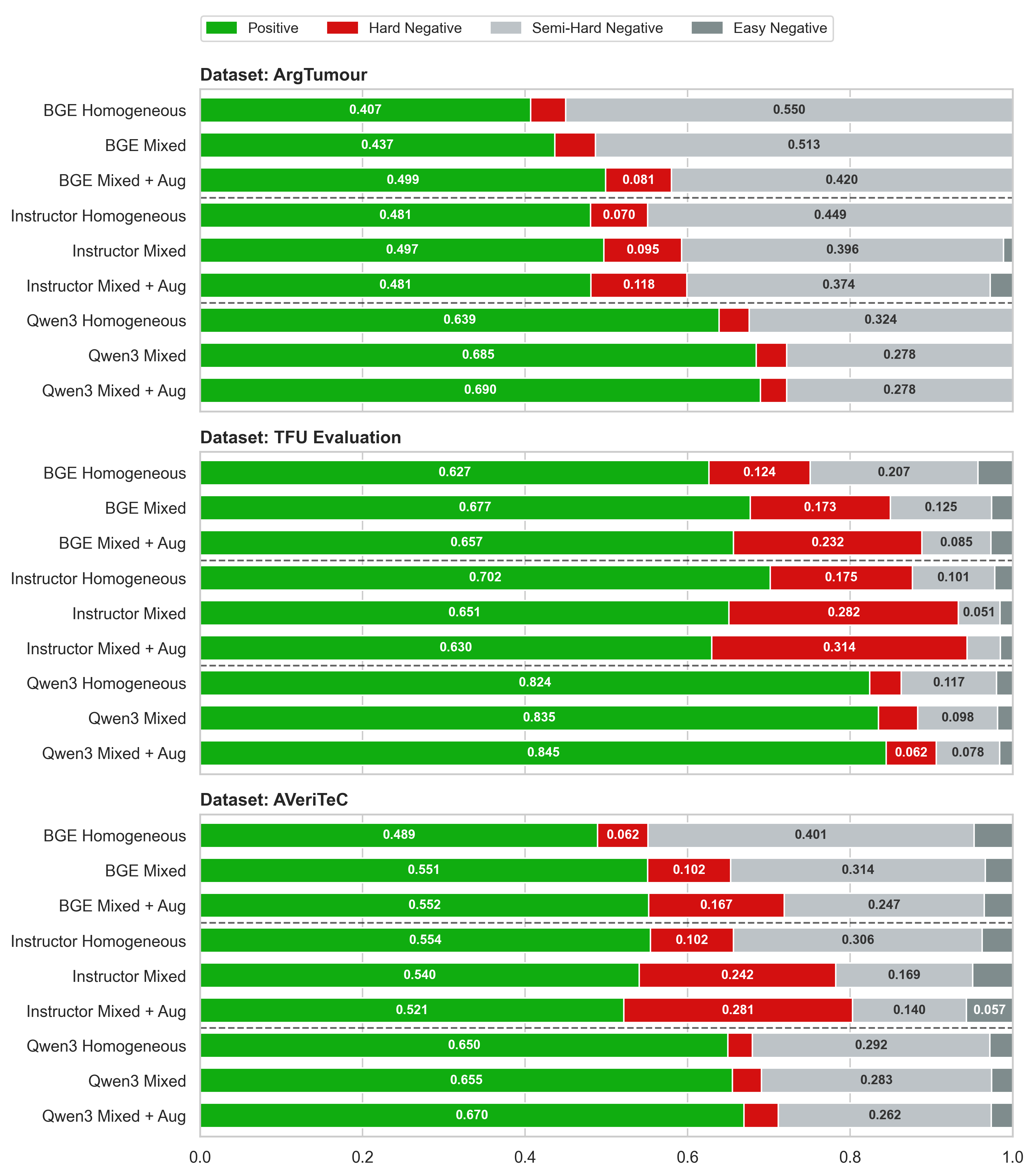}
    \caption{Retrieval composition across the ArgTumour, TFU Evaluation, and AVeriTeC datasets. The chart contrasts the homogeneous baseline against the data-centric interventions (Mixed and Mixed + Aug) across the three model architectures. The balanced curriculum (Mixed) mitigates topical collapse by uniformly reducing the proportion of semi-hard negatives (light grey segments) across all models. However, architectural responses diverge significantly: while Qwen3-Embedding-8B slightly expands its positive retrieval (green segments), both BGE-Large and Instructor-XL exhibit a severe stance regression (inflated red segments) when the hard-negative pressure is reduced.  More detailed results can be found in Appendix \ref{app:retrieval_results}}
    \label{fig:chapter_6_retrieval}
\end{figure}

To evaluate the impact of the proposed interventions, we compare the retrieval metrics across the base models, the homogeneous fine-tuned baselines from Section \ref{sec:baseline_model_evaluation}, and models using our interventions. The resulting retrieval compositions (visualised in Figure \ref{fig:chapter_6_retrieval}) and metrics reveal substantial differences:
\begin{itemize}
\item \textbf{BGE-Large}: Results reveal that BGE-Large struggles to balance topic and stance. While the homogeneous curriculum achieved a Precision@R of 0.627 on the TFU Evaluation dataset, our interventions traded stance sensitivity for topical focus. Applying the Mixed + Aug curriculum successfully reduced semi-hard negatives (from 20.7\% to 8.5\%), but triggered a noticeable regression in relational logic. This nearly doubled the hard negative rate to 23.2\%, suggesting that BGE-Large lacks the capacity to robustly maintain both boundaries simultaneously.
\item \textbf{Instructor-XL}: Instructor-XL demonstrated a strong dependency on high-density hard-negative signals. In its base state on the TFU Evaluation dataset, it was largely stance-blind with a hard-negative rate of 41.1\%. While the homogeneous curriculum reduced this down to 17.5\%, it also triggered topical collapse, increasing the semi-hard negative rate from 2.7\% to 10.1\%. When the `Mixed' curriculum was applied to restore topic-sensitivity, the semi-hard negative rate successfully dropped to 5.1\%, but the model seemingly lacked the capacity to robustly learn relational logic. Consequently, its stance error rate regressed to 31.4\%.
\item \textbf{Qwen3-Embedding-8B}: The decoder-only model emerged as the architecture most capable in stance-topic adaptation. The homogeneous curriculum fixed its base stance blindness (reducing stance error from 33.5\% to 3.9\% on TFU Evaluation) but introduced the expected topical collapse. However, fine-tuning on the `Mixed' curriculum successfully reduced the semi-hard negative rate from 11.7\% to 9.8\%, while maintaining good stance sensitivity. Furthermore, introducing the synthetic stance inversions (`Mixed + Aug') further reduced this rate to 7.8\%. This resulted in overall best performance, with a Precision@R of 0.845 on the TFU Evaluation dataset.
\end{itemize}

Finally, evaluating the hybrid fusion (dense `Mixed + Aug' combined with BM25 via RSF) confirmed our hypothesis regarding the division of responsibility. The addition of the sparse retriever provided a topical safety net for specialised domains, improving Qwen3-Embedding-8B's Precision@R on the ArgTumour dataset from 0.690 up to 0.723 (at $\lambda=0.1$). However, because BM25 is inherently stance-blind, this lexical boost came at the direct cost of inflating the stance error across all models as the sparse weight ($\lambda$) increased. Consequently, hybrid fusion proved highly effective for dense, domain-specific corpora (ArgTumour, AVeriTeC) where entity matching is critical, but offered diminishing returns on general-domain datasets (TFU Evaluation) where the unhindered stance logic of the tuned bi-encoder was already sufficient. Results capturing hybrid retrieval performance are provided in Figure \ref{fig:hybrid_search_results}.

\subsection{Word Ablation Validation}
\begin{figure}
    \centering
    \includegraphics[width=1\linewidth]{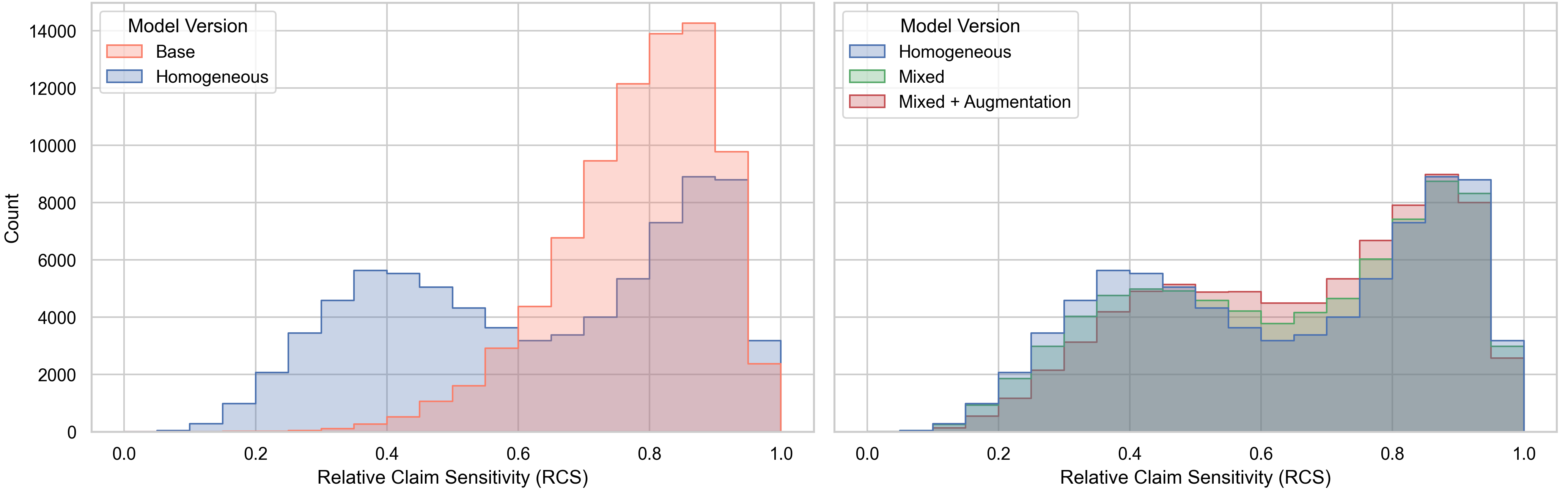}
    \caption{Relative Claim Sensitivity (RCS) distributions for Qwen3-Embedding-8B on AVeriTeC data. \textbf{Left}: The Base model exhibits a severe noun bias, allocating the vast majority of its representational capacity to the claim (red distribution). Applying the homogeneous curriculum triggers topical collapse, shifting focus away from the claim and creating a notable secondary peak with low claim sensitivity (blue distribution). \textbf{Right}: The data-centric interventions regularise the model's instruction focus. The Mixed (green) and Mixed + Augmentation (red) curricula progressively flatten the peak and pull semantic focus back toward the underlying topic.}
\label{fig:rcs_ablation}
\end{figure}

To diagnose the shifting retrieval distributions observed in Section \ref{subsec:intervention_evaluation}, we again applied the word ablation metrics. Figure \ref{fig:rcs_ablation} visualises the Relative Claim Sensitivity (RCS) for Qwen on the AVeriTeC dataset. The base model exhibited a severe topic bias, allocating a vast majority of its representational capacity to the claim, resulting in a mean RCS of $0.783$. Applying the homogeneous curriculum resulted in topical collapse: the mean RCS dropped sharply to $0.637$, creating a notable secondary density peak as the model representations became less sensitive to the topic compared to the instruction.

The application of the data-centric interventions counteracted this phenomenon. By introducing the `Mixed + Aug' curriculum, the collapse peak was flattened, and semantic focus partially shifted back toward the claim, causing the mean RCS to increase to $0.661$. Crucially, evaluating the model's Directional Impact (DI) confirmed that this recovered topical sensitivity did not come at the cost of relational logic. While the base model practically ignored stance verbs (DI for hard negatives: $0.024$), the `Mixed + Aug' model maintained a strong, asymmetric stance boundary (DI for hard negatives: $0.144$, DI for positives: $-0.120$). Computing the ablation metrics across the broader evaluation suite confirmed these general trends across all datasets. Full ablation metric results are given in Appendix \ref{app:ablation_metrics_tables}.

\section{Conclusion}

In this paper, we diagnosed and addressed the inability of base embedding models to balance topical subject matter with directional stance during argument retrieval. Baseline evaluations confirmed that models initially exhibit a severe topic bias, disregarding stance constraints in the instructions. While contrastive fine-tuning on a homogeneous curriculum of hard negatives improves stance sensitivity, it can also degrade the ability of models to correctly represent the topic, resulting in \textit{topical collapse}.

To resolve this collapse, we proposed a two-stage data-centric pipeline. A balanced curriculum increased topical sensitivity across all evaluated architectures, while synthetic stance inversions reduced residual lexical shortcuts. This approach proved particularly effective for Qwen3-Embedding-8B, improving its ability to jointly encode stance and topical relevance as well as the overall retrieval performance. Furthermore, integrating a BM25 sparse retriever provided an additional topical safety net, patching the dense model's remaining blind spots in highly specialised domains.

Future work could further explore applications of our tuned models to argument mining from large text corpora or for argumentative fact verification (e.g., in the spirit of \cite{freedman-2025-argumentative-llms,zhu-2025-argrag}). Additionally, while our data-centric paradigm was reasonably effective at counteracting topical collapse, future research could explore model-centric interventions, such as attention steering, to further improve performance. Finally, it would be interesting to explore dynamically adapting the weighting factor $\lambda$ in our hybrid search pipeline to specific queries or corpora.

\begin{acknowledgments}
Dejl and Toni were partially funded by the European Research Council (ERC) under the European Union’s Horizon 2020 research and innovation programme (grant agreement No. 101020934, ADIX). Toni was also funded by EPSRC (grant UKRI3928, NeSyDebates). 
\end{acknowledgments}

\section*{Declaration on Generative AI}
The authors used Gemini 3 Pro to generate standardised training/evaluation instructions and for code assistance (boilerplate PyTorch code, code generating Matplotlib figures and LaTeX tables). Further, the authors used Gemini 3.5 Flash to generate synthetic, stance-inverted arguments for dataset augmentation. Finally, GPT-5.5 was used to provide general feedback on the draft manuscript. After using these tools, the authors reviewed and edited the outputs as needed and take full responsibility for the paper content.

\bibliography{references}

\appendix

\section{Appendices Outline}
The appendices provide supplementary materials, detailed experimental results, and extended evaluations to support the findings in the main text. Appendix \ref{app:instr_lexicon} details the complete set of seen and unseen instructions utilised for the stance-aware retrieval tasks, and Appendix \ref{app:experimental_details} provides the experimental details. Appendices \ref{app:retrieval_results} and \ref{app:ablation_metrics_tables} supply comprehensive tabular breakdowns of the retrieval performance and word-ablation metrics across all evaluated models and datasets. Subsequent sections offer expanded visual diagnostics, including supplementary Relative Claim Sensitivity (RCS) and Directional Impact (DI) plots in Appendix \ref{app:ablation_metrics_figs}, as well as detailed word-ablation heatmaps and their corresponding argument texts in Appendix \ref{app:ablation_heatmap}. Finally, Appendix \ref{app:stance_inversion_prompt} provides the exact one-shot prompt utilised for synthetic data generation, and Appendix \ref{app:hybrid_retrieval_results} visualises the retrieval results of the hybrid retrieval pipeline. 

\section{Instruction Sets}
\label{app:instr_lexicon}

Overview of the seen/unseen (i.e., training/testing) instructions used for support/attack stance retrieval tasks respectively.

\section*{Seen Instructions}

\subsection*{Support}
\begin{itemize}
    \item Find evidence backing
    \item Retrieve arguments in favor of
    \item Search for statements validating
    \item Highlight information corroborating
    \item Uncover points upholding
    \item Show claims affirming
    \item Provide reasoning that supports
    \item Identify excerpts substantiating
    \item Extract statements advocating for
    \item Gather facts verifying
\end{itemize}

\subsection*{Attack}
\begin{itemize}
    \item Find evidence refuting
    \item Retrieve arguments against
    \item Search for statements invalidating
    \item Highlight information contradicting
    \item Uncover points opposing
    \item Show claims challenging
    \item Provide reasoning that undermines
    \item Identify excerpts disputing
    \item Extract statements arguing against
    \item Gather facts debunking
\end{itemize}

\section*{Unseen Instructions}

\subsection*{Support}
\begin{itemize}
    \item Discover sources promoting
    \item Retrieve evidence confirming
    \item Locate specific arguments justifying
    \item Find statements endorsing
    \item Show points backing up
\end{itemize}

\subsection*{Attack}
\begin{itemize}
    \item Discover sources dismissing
    \item Retrieve evidence disproving
    \item Locate specific arguments attacking
    \item Find statements rejecting
    \item Show points doubting
\end{itemize}

\clearpage
\section{Experimental Details}
\label{app:experimental_details}

This section outlines the hardware, hyperparameters, and dataset statistics utilised during our experiments.

\paragraph{Hardware} Model training and inference were distributed across two high-performance compute clusters. Experiments utilised L40S GPUs hosted on the Imperial College HPC Cluster, as well as A40 GPUs hosted on the Department of Computing (Doc) GPU Cluster.

\paragraph{Training Hyperparameters} All fine-tuning experiments were optimised using Multiple Negatives Ranking Loss (MNRL) \cite{henderson2017efficientnaturallanguageresponse}. The models were trained for 2 epochs with a per-device batch size of 8 and a gradient accumulation step of 2. The optimisation process utilised a learning rate of $2 \times 10^{-5}$ with a warmup ratio of 0.1. Depending on the architecture, training utilised either FP16 or BF16 precision. For random seeds, the training code relied on the standard default initialisation provided by the Transformers library, with dynamic claim shuffling handled at the batch-sampler level.

\paragraph{LoRA Configuration} For fine-tuning via Low-Rank Adaption (LoRA), the $\alpha$ scaling parameter was dynamically set to $2 \times r$ (where $r = 64$), alongside a dropout rate of $0.05$. The target modules varied by architecture:
\begin{itemize}
    \item \textbf{BGE-Large:} \texttt{query}, \texttt{key}, \texttt{value}, \texttt{dense}
    \item \textbf{Instructor-XL:} \texttt{q}, \texttt{k}, \texttt{v}, \texttt{o}
    \item \textbf{Qwen3-Embedding-8B:} \texttt{q\_proj}, \texttt{k\_proj}, \texttt{v\_proj}, \texttt{o\_proj}, \texttt{gate\_proj}, \texttt{up\_proj}, \texttt{down\_proj}
\end{itemize}

\paragraph{Dataset Statistics} Table \ref{tab:dataset_statistics} provides a comprehensive breakdown of the datasets used for validation and zero-shot evaluation.

\begin{table*}[htbp]
\centering
\caption{Dataset statistics detailing the number of claims, total supporting/attacking arguments, and the average number of supporting/attacking arguments per claim (with minimum and maximum bounds).}
\label{tab:dataset_statistics}
\renewcommand{\arraystretch}{1.2}
\resizebox{\textwidth}{!}{%
\begin{tabular}{l c cccc}
\toprule[1.5pt]
\textbf{Dataset} & \textbf{Claims} & \textbf{Total Sup.} & \textbf{Avg Sup. (Min-Max)} & \textbf{Total Att.} & \textbf{Avg Att. (Min-Max)} \\
\midrule[1.5pt]
TFU Training Arguments (80\% for training) & 283 & 1051 & 3.71 (0-7) & 1146 & 4.05 (0-7) \\
TFU Training Arguments (20\% held-out) & 71 & 295 & 4.15 (1-6) & 303 & 4.27 (2-6) \\
TFU Validation Arguments (GPT-5) & 150 & 622 & 4.15 (0-12) & 882 & 5.88 (0-13) \\
TFU Validation Arguments (Qwen3-8B) & 150 & 374 & 2.49 (0-5) & 485 & 3.23 (0-5) \\
ArgTumour Arguments & 14 & 40 & 2.86 (0-16) & 58 & 4.14 (2-10) \\
TFU Evaluation Arguments & 750 & 1819 & 2.43 (0-5) & 2321 & 3.09 (0-5) \\
AVeriTeC Arguments & 1746 & 3180 & 1.82 (0-5) & 5310 & 3.04 (0-5) \\
\bottomrule[1.5pt]
\end{tabular}%
}
\end{table*}

\clearpage

\section{Retrieval Metrics Tables}
\label{app:retrieval_results}

The following tables present comprehensive stance retrieval metrics across all evaluated datasets and model families. Results represent the mean ($\mu$) and standard deviation ($\sigma$). Hard@R, Semi@R, and Easy@R denote the proportion of retrieved hard, semi-hard, and easy negatives within the top $R$ results, respectively. The absolute best score (highest for NDCG/Precision, lowest for error rates) is highlighted in \textbf{bold}, while the best score within each dense/hybrid subgroup is \underline{underlined}. Note that all Hybrid Fusion (RSF) strategies operate exclusively over the Mixed + Aug base models. Note that the reported standard deviations ($\sigma$) reflect the variance across different evaluation queries and \textit{unseen} instructions.

\vspace{2em}

\begin{table*}[htbp]
\centering
\caption{Retrieval performance on the \textbf{TFU Training Arguments (20\% held-out)} dataset using \textbf{BGE-Large}.}
\label{tab:results_tfu_training_arguments_20_held-out_bge_large}
\renewcommand{\arraystretch}{1.2}
\begin{tabular}{l c ccccc}
\toprule[1.5pt]
\textbf{Strategy} & \textbf{Stance} & \textbf{NDCG@10} & \textbf{Prec@R} & \textbf{Hard@R} & \textbf{Semi@R} & \textbf{Easy@R} \\
\midrule[1.5pt]
\multicolumn{7}{l}{\textbf{GROUP 1: Dense Retrieval}} \\
\multirow{3}{*}{Zero-Shot} & Attack & $0.613\pm0.28$ & $0.438\pm0.27$ & $0.342\pm0.26$ & $0.181\pm0.27$ & $0.038\pm0.12$ \\
 & Support & $0.605\pm0.27$ & $0.401\pm0.27$ & $0.420\pm0.25$ & $0.043\pm0.12$ & $0.136\pm0.23$ \\
 & Overall & $0.609\pm0.27$ & $0.419\pm0.27$ & $0.381\pm0.26$ & $0.112\pm0.22$ & $0.087\pm0.19$ \\
\addlinespace[1.5em]
\multirow{3}{*}{Homogeneous} & Attack & $0.883\pm0.22$ & $0.795\pm0.27$ & $\underline{\mathbf{0.087\pm0.17}}$ & $0.112\pm0.19$ & $0.006\pm0.03$ \\
 & Support & $\underline{\mathbf{0.936\pm0.12}}$ & $\underline{\mathbf{0.850\pm0.20}}$ & $\underline{\mathbf{0.088\pm0.16}}$ & $0.053\pm0.13$ & $0.010\pm0.05$ \\
 & Overall & $0.909\pm0.18$ & $0.822\pm0.24$ & $\underline{\mathbf{0.088\pm0.17}}$ & $0.082\pm0.17$ & $0.008\pm0.04$ \\
\end{tabular}
\end{table*}

\vspace{3em}

\begin{table*}[htbp]
\centering
\caption{Retrieval performance on the \textbf{TFU Training Arguments (20\% held-out)} dataset using \textbf{Instructor-XL}.}
\label{tab:results_tfu_training_arguments_20_held-out_instructor_xl}
\renewcommand{\arraystretch}{1.2}
\begin{tabular}{l c ccccc}
\toprule[1.5pt]
\textbf{Strategy} & \textbf{Stance} & \textbf{NDCG@10} & \textbf{Prec@R} & \textbf{Hard@R} & \textbf{Semi@R} & \textbf{Easy@R} \\
\midrule[1.5pt]
\multicolumn{7}{l}{\textbf{GROUP 1: Dense Retrieval}} \\
\multirow{3}{*}{Zero-Shot} & Attack & $0.725\pm0.14$ & $0.482\pm0.19$ & $0.507\pm0.20$ & $\underline{\mathbf{0.008\pm0.06}}$ & $0.003\pm0.02$ \\
 & Support & $0.804\pm0.14$ & $0.581\pm0.19$ & $0.410\pm0.19$ & $0.006\pm0.04$ & $\underline{0.003\pm0.03}$ \\
 & Overall & $0.764\pm0.15$ & $0.531\pm0.20$ & $0.459\pm0.20$ & $\underline{\mathbf{0.007\pm0.05}}$ & $\underline{0.003\pm0.02}$ \\
\addlinespace[1.5em]
\multirow{3}{*}{Homogeneous} & Attack & $\underline{\mathbf{0.872\pm0.23}}$ & $\underline{\mathbf{0.773\pm0.29}}$ & $\underline{\mathbf{0.139\pm0.24}}$ & $0.080\pm0.17$ & $0.009\pm0.04$ \\
 & Support & $\underline{\mathbf{0.965\pm0.09}}$ & $\underline{\mathbf{0.905\pm0.17}}$ & $\underline{\mathbf{0.049\pm0.12}}$ & $0.030\pm0.08$ & $0.015\pm0.07$ \\
 & Overall & $\underline{\mathbf{0.919\pm0.18}}$ & $\underline{\mathbf{0.839\pm0.24}}$ & $\underline{\mathbf{0.094\pm0.20}}$ & $0.055\pm0.13$ & $0.012\pm0.06$ \\
\end{tabular}
\end{table*}

\vspace{3em}

\begin{table*}[htbp]
\centering
\caption{Retrieval performance on the \textbf{TFU Training Arguments (20\% held-out)} dataset using \textbf{Qwen3-Embedding-8B}.}
\label{tab:results_tfu_training_arguments_20_held-out_qwen3_embedding_8b}
\renewcommand{\arraystretch}{1.2}
\begin{tabular}{l c ccccc}
\toprule[1.5pt]
\textbf{Strategy} & \textbf{Stance} & \textbf{NDCG@10} & \textbf{Prec@R} & \textbf{Hard@R} & \textbf{Semi@R} & \textbf{Easy@R} \\
\midrule[1.5pt]
\multicolumn{7}{l}{\textbf{GROUP 1: Dense Retrieval}} \\
\multirow{3}{*}{Zero-Shot} & Attack & $0.792\pm0.15$ & $0.559\pm0.22$ & $0.430\pm0.22$ & $0.006\pm0.03$ & $0.004\pm0.03$ \\
 & Support & $0.866\pm0.13$ & $0.634\pm0.21$ & $0.363\pm0.21$ & $0.003\pm0.02$ & $\underline{\mathbf{0.000\pm0.00}}$ \\
 & Overall & $0.829\pm0.14$ & $0.597\pm0.21$ & $0.397\pm0.22$ & $0.005\pm0.03$ & $0.002\pm0.02$ \\
\addlinespace[1.5em]
\multirow{3}{*}{Zero-Shot + Reranker} & Attack & $0.779\pm0.16$ & $0.535\pm0.23$ & $0.465\pm0.24$ & $\underline{\mathbf{0.000\pm0.00}}$ & $\underline{\mathbf{0.001\pm0.01}}$ \\
 & Support & $0.874\pm0.13$ & $0.695\pm0.22$ & $0.305\pm0.22$ & $\underline{\mathbf{0.000\pm0.00}}$ & $\underline{\mathbf{0.000\pm0.00}}$ \\
 & Overall & $0.826\pm0.15$ & $0.615\pm0.24$ & $0.385\pm0.24$ & $\underline{\mathbf{0.000\pm0.00}}$ & $\underline{\mathbf{0.000\pm0.01}}$ \\
\end{tabular}
\end{table*}

\vspace{3em}

\begin{table*}[htbp]
\centering
\caption{Retrieval performance on the \textbf{TFU Validation Arguments generated by GPT-5} dataset using \textbf{BGE-Large}.}
\label{tab:results_tfu_validation_arguments_generated_by_gpt-5_bge_large}
\renewcommand{\arraystretch}{1.2}
\begin{tabular}{l c ccccc}
\toprule[1.5pt]
\textbf{Strategy} & \textbf{Stance} & \textbf{NDCG@10} & \textbf{Prec@R} & \textbf{Hard@R} & \textbf{Semi@R} & \textbf{Easy@R} \\
\midrule[1.5pt]
\multicolumn{7}{l}{\textbf{GROUP 1: Dense Retrieval}} \\
\multirow{3}{*}{Zero-Shot} & Attack & $0.562\pm0.31$ & $0.448\pm0.30$ & $0.314\pm0.28$ & $\underline{\mathbf{0.131\pm0.18}}$ & $0.106\pm0.15$ \\
 & Support & $0.678\pm0.22$ & $0.523\pm0.21$ & $0.261\pm0.19$ & $\underline{0.120\pm0.18}$ & $0.096\pm0.15$ \\
 & Overall & $0.610\pm0.28$ & $0.480\pm0.27$ & $0.292\pm0.25$ & $\underline{0.126\pm0.18}$ & $0.102\pm0.15$ \\
\addlinespace[1.5em]
\multirow{3}{*}{Homogeneous} & Attack & $0.761\pm0.25$ & $0.655\pm0.27$ & $\underline{\mathbf{0.022\pm0.07}}$ & $0.317\pm0.26$ & $\underline{\mathbf{0.006\pm0.03}}$ \\
 & Support & $0.822\pm0.19$ & $0.730\pm0.24$ & $\underline{\mathbf{0.046\pm0.11}}$ & $0.207\pm0.22$ & $0.017\pm0.07$ \\
 & Overall & $0.787\pm0.23$ & $0.686\pm0.26$ & $\underline{\mathbf{0.032\pm0.09}}$ & $0.271\pm0.25$ & $0.011\pm0.05$ \\
\end{tabular}
\end{table*}

\vspace{3em}

\begin{table*}[htbp]
\centering
\caption{Retrieval performance on the \textbf{TFU Validation Arguments generated by GPT-5} dataset using \textbf{Instructor-XL}.}
\label{tab:results_tfu_validation_arguments_generated_by_gpt-5_instructor_xl}
\renewcommand{\arraystretch}{1.2}
\begin{tabular}{l c ccccc}
\toprule[1.5pt]
\textbf{Strategy} & \textbf{Stance} & \textbf{NDCG@10} & \textbf{Prec@R} & \textbf{Hard@R} & \textbf{Semi@R} & \textbf{Easy@R} \\
\midrule[1.5pt]
\multicolumn{7}{l}{\textbf{GROUP 1: Dense Retrieval}} \\
\multirow{3}{*}{Zero-Shot} & Attack & $0.662\pm0.26$ & $0.529\pm0.32$ & $0.336\pm0.31$ & $\underline{\mathbf{0.079\pm0.14}}$ & $0.056\pm0.13$ \\
 & Support & $0.694\pm0.24$ & $0.558\pm0.23$ & $0.319\pm0.20$ & $\underline{\mathbf{0.070\pm0.15}}$ & $0.054\pm0.11$ \\
 & Overall & $0.675\pm0.25$ & $0.541\pm0.29$ & $0.328\pm0.27$ & $\underline{\mathbf{0.075\pm0.15}}$ & $0.055\pm0.12$ \\
\addlinespace[1.5em]
\multirow{3}{*}{Homogeneous} & Attack & $\underline{\mathbf{0.786\pm0.24}}$ & $\underline{\mathbf{0.687\pm0.28}}$ & $\underline{\mathbf{0.043\pm0.13}}$ & $0.263\pm0.26$ & $\underline{\mathbf{0.007\pm0.05}}$ \\
 & Support & $\underline{\mathbf{0.841\pm0.19}}$ & $\underline{\mathbf{0.754\pm0.23}}$ & $\underline{\mathbf{0.048\pm0.10}}$ & $0.177\pm0.21$ & $0.021\pm0.06$ \\
 & Overall & $\underline{\mathbf{0.809\pm0.23}}$ & $\underline{\mathbf{0.715\pm0.26}}$ & $\underline{\mathbf{0.045\pm0.12}}$ & $0.227\pm0.25$ & $\underline{\mathbf{0.013\pm0.06}}$ \\
\end{tabular}
\end{table*}

\vspace{3em}

\begin{table*}[htbp]
\centering
\caption{Retrieval performance on the \textbf{TFU Validation Arguments generated by GPT-5} dataset using \textbf{Qwen3-Embedding-8B}.}
\label{tab:results_tfu_validation_arguments_generated_by_gpt-5_qwen3_embedding_8b}
\renewcommand{\arraystretch}{1.2}
\begin{tabular}{l c ccccc}
\toprule[1.5pt]
\textbf{Strategy} & \textbf{Stance} & \textbf{NDCG@10} & \textbf{Prec@R} & \textbf{Hard@R} & \textbf{Semi@R} & \textbf{Easy@R} \\
\midrule[1.5pt]
\multicolumn{7}{l}{\textbf{GROUP 1: Dense Retrieval}} \\
\multirow{3}{*}{Zero-Shot} & Attack & $0.730\pm0.24$ & $0.600\pm0.31$ & $0.314\pm0.29$ & $\underline{\mathbf{0.057\pm0.13}}$ & $0.030\pm0.10$ \\
 & Support & $0.803\pm0.18$ & $0.667\pm0.21$ & $0.256\pm0.19$ & $\underline{\mathbf{0.055\pm0.15}}$ & $0.022\pm0.06$ \\
 & Overall & $0.760\pm0.22$ & $0.628\pm0.27$ & $0.289\pm0.26$ & $\underline{\mathbf{0.056\pm0.14}}$ & $0.027\pm0.09$ \\
\addlinespace[1.5em]
\multirow{3}{*}{Homogeneous} & Attack & $0.906\pm0.14$ & $0.828\pm0.19$ & $\underline{\mathbf{0.012\pm0.06}}$ & $0.154\pm0.19$ & $0.006\pm0.04$ \\
 & Support & $0.926\pm0.13$ & $0.877\pm0.19$ & $\underline{\mathbf{0.020\pm0.06}}$ & $0.095\pm0.18$ & $0.008\pm0.03$ \\
 & Overall & $0.914\pm0.14$ & $0.849\pm0.19$ & $\underline{\mathbf{0.015\pm0.06}}$ & $0.129\pm0.19$ & $0.007\pm0.04$ \\
\end{tabular}
\end{table*}

\vspace{3em}

\begin{table*}[htbp]
\centering
\caption{Retrieval performance on the \textbf{TFU Validation Arguments generated by Qwen3-8B} dataset using \textbf{BGE-Large}.}
\label{tab:results_tfu_validation_arguments_generated_by_qwen3-8b_bge_large}
\renewcommand{\arraystretch}{1.2}
\begin{tabular}{l c ccccc}
\toprule[1.5pt]
\textbf{Strategy} & \textbf{Stance} & \textbf{NDCG@10} & \textbf{Prec@R} & \textbf{Hard@R} & \textbf{Semi@R} & \textbf{Easy@R} \\
\midrule[1.5pt]
\multicolumn{7}{l}{\textbf{GROUP 1: Dense Retrieval}} \\
\multirow{3}{*}{Zero-Shot} & Attack & $0.619\pm0.24$ & $0.408\pm0.31$ & $0.481\pm0.30$ & $\underline{\mathbf{0.036\pm0.11}}$ & $0.075\pm0.16$ \\
 & Support & $0.823\pm0.22$ & $0.611\pm0.29$ & $0.296\pm0.29$ & $\underline{0.067\pm0.16}$ & $0.026\pm0.09$ \\
 & Overall & $0.715\pm0.25$ & $0.503\pm0.32$ & $0.394\pm0.31$ & $\underline{\mathbf{0.051\pm0.14}}$ & $0.052\pm0.13$ \\
\addlinespace[1.5em]
\multirow{3}{*}{Homogeneous} & Attack & $0.807\pm0.22$ & $0.658\pm0.26$ & $\underline{\mathbf{0.064\pm0.14}}$ & $0.270\pm0.27$ & $\underline{\mathbf{0.008\pm0.05}}$ \\
 & Support & $0.844\pm0.24$ & $0.734\pm0.33$ & $\underline{\mathbf{0.104\pm0.23}}$ & $0.134\pm0.26$ & $0.028\pm0.12$ \\
 & Overall & $0.824\pm0.23$ & $0.694\pm0.30$ & $\underline{\mathbf{0.083\pm0.19}}$ & $0.206\pm0.27$ & $0.017\pm0.09$ \\
\end{tabular}
\end{table*}

\vspace{3em}

\begin{table*}[htbp]
\centering
\caption{Retrieval performance on the \textbf{TFU Validation Arguments generated by Qwen3-8B} dataset using \textbf{Instructor-XL}.}
\label{tab:results_tfu_validation_arguments_generated_by_qwen3-8b_instructor_xl}
\renewcommand{\arraystretch}{1.2}
\begin{tabular}{l c ccccc}
\toprule[1.5pt]
\textbf{Strategy} & \textbf{Stance} & \textbf{NDCG@10} & \textbf{Prec@R} & \textbf{Hard@R} & \textbf{Semi@R} & \textbf{Easy@R} \\
\midrule[1.5pt]
\multicolumn{7}{l}{\textbf{GROUP 1: Dense Retrieval}} \\
\multirow{3}{*}{Zero-Shot} & Attack & $0.710\pm0.22$ & $0.492\pm0.33$ & $0.445\pm0.30$ & $\underline{\mathbf{0.025\pm0.09}}$ & $0.038\pm0.12$ \\
 & Support & $0.820\pm0.21$ & $0.595\pm0.30$ & $0.342\pm0.31$ & $\underline{0.041\pm0.13}$ & $0.022\pm0.09$ \\
 & Overall & $0.762\pm0.22$ & $0.540\pm0.32$ & $0.397\pm0.31$ & $\underline{\mathbf{0.033\pm0.11}}$ & $0.030\pm0.11$ \\
\addlinespace[1.5em]
\multirow{3}{*}{Homogeneous} & Attack & $\underline{\mathbf{0.844\pm0.19}}$ & $\underline{\mathbf{0.687\pm0.26}}$ & $\underline{\mathbf{0.100\pm0.16}}$ & $0.209\pm0.25$ & $\underline{\mathbf{0.004\pm0.04}}$ \\
 & Support & $\underline{\mathbf{0.895\pm0.21}}$ & $\underline{\mathbf{0.783\pm0.31}}$ & $\underline{\mathbf{0.103\pm0.23}}$ & $0.106\pm0.24$ & $\underline{\mathbf{0.008\pm0.06}}$ \\
 & Overall & $\underline{\mathbf{0.868\pm0.20}}$ & $\underline{\mathbf{0.732\pm0.29}}$ & $\underline{\mathbf{0.101\pm0.20}}$ & $0.161\pm0.25$ & $\underline{\mathbf{0.006\pm0.05}}$ \\
\end{tabular}
\end{table*}

\vspace{3em}

\begin{table*}[htbp]
\centering
\caption{Retrieval performance on the \textbf{TFU Validation Arguments generated by Qwen3-8B} dataset using \textbf{Qwen3-Embedding-8B}.}
\label{tab:results_tfu_validation_arguments_generated_by_qwen3-8b_qwen3_embedding_8b}
\renewcommand{\arraystretch}{1.2}
\begin{tabular}{l c ccccc}
\toprule[1.5pt]
\textbf{Strategy} & \textbf{Stance} & \textbf{NDCG@10} & \textbf{Prec@R} & \textbf{Hard@R} & \textbf{Semi@R} & \textbf{Easy@R} \\
\midrule[1.5pt]
\multicolumn{7}{l}{\textbf{GROUP 1: Dense Retrieval}} \\
\multirow{3}{*}{Zero-Shot} & Attack & $0.780\pm0.20$ & $0.566\pm0.30$ & $0.387\pm0.28$ & $\underline{\mathbf{0.027\pm0.09}}$ & $0.020\pm0.09$ \\
 & Support & $0.881\pm0.20$ & $0.705\pm0.31$ & $0.242\pm0.31$ & $\underline{\mathbf{0.038\pm0.11}}$ & $0.015\pm0.08$ \\
 & Overall & $0.828\pm0.20$ & $0.631\pm0.31$ & $0.319\pm0.30$ & $\underline{\mathbf{0.032\pm0.10}}$ & $0.018\pm0.08$ \\
\addlinespace[1.5em]
\multirow{3}{*}{Homogeneous} & Attack & $0.932\pm0.12$ & $0.832\pm0.21$ & $0.025\pm0.10$ & $0.136\pm0.20$ & $0.007\pm0.05$ \\
 & Support & $0.903\pm0.25$ & $0.836\pm0.30$ & $\underline{\mathbf{0.024\pm0.12}}$ & $0.125\pm0.27$ & $0.014\pm0.09$ \\
 & Overall & $0.918\pm0.19$ & $0.834\pm0.26$ & $\underline{\mathbf{0.025\pm0.11}}$ & $0.131\pm0.24$ & $0.010\pm0.07$ \\
\end{tabular}
\end{table*}

\vspace{3em}

\begin{table*}[htbp]
\centering
\caption{Retrieval performance on the \textbf{ArgTumour Arguments} dataset using \textbf{BGE-Large}.}
\label{tab:results_argtumour_arguments_bge_large}
\renewcommand{\arraystretch}{1.2}
\begin{tabular}{l c ccccc}
\toprule[1.5pt]
\textbf{Strategy} & \textbf{Stance} & \textbf{NDCG@10} & \textbf{Prec@R} & \textbf{Hard@R} & \textbf{Semi@R} & \textbf{Easy@R} \\
\midrule[1.5pt]
\multicolumn{7}{l}{\textbf{GROUP 1: Dense Retrieval}} \\
\multirow{3}{*}{Zero-Shot} & Attack & $0.485\pm0.40$ & $0.317\pm0.32$ & $0.128\pm0.25$ & $0.513\pm0.31$ & $0.043\pm0.09$ \\
 & Support & $0.338\pm0.30$ & $0.259\pm0.29$ & $0.089\pm0.13$ & $\underline{\mathbf{0.381\pm0.12}}$ & $0.271\pm0.15$ \\
 & Overall & $0.441\pm0.38$ & $0.299\pm0.31$ & $0.116\pm0.22$ & $0.473\pm0.27$ & $0.111\pm0.15$ \\
\addlinespace[1.5em]
\multirow{3}{*}{Homogeneous} & Attack & $0.567\pm0.34$ & $0.444\pm0.31$ & $\underline{\mathbf{0.021\pm0.08}}$ & $0.535\pm0.30$ & $\underline{\mathbf{0.000\pm0.00}}$ \\
 & Support & $0.421\pm0.32$ & $\underline{0.321\pm0.25}$ & $0.094\pm0.14$ & $0.585\pm0.23$ & $\underline{\mathbf{0.000\pm0.00}}$ \\
 & Overall & $0.523\pm0.34$ & $0.407\pm0.30$ & $\underline{\mathbf{0.043\pm0.11}}$ & $0.550\pm0.28$ & $\underline{\mathbf{0.000\pm0.00}}$ \\
\end{tabular}
\end{table*}

\vspace{3em}

\begin{table*}[htbp]
\centering
\caption{Retrieval performance on the \textbf{ArgTumour Arguments} dataset using \textbf{Instructor-XL}.}
\label{tab:results_argtumour_arguments_instructor_xl}
\renewcommand{\arraystretch}{1.2}
\begin{tabular}{l c ccccc}
\toprule[1.5pt]
\textbf{Strategy} & \textbf{Stance} & \textbf{NDCG@10} & \textbf{Prec@R} & \textbf{Hard@R} & \textbf{Semi@R} & \textbf{Easy@R} \\
\midrule[1.5pt]
\multicolumn{7}{l}{\textbf{GROUP 1: Dense Retrieval}} \\
\multirow{3}{*}{Zero-Shot} & Attack & $0.579\pm0.38$ & $0.503\pm0.39$ & $0.157\pm0.26$ & $\underline{0.287\pm0.30}$ & $0.054\pm0.10$ \\
 & Support & $0.473\pm0.25$ & $0.371\pm0.27$ & $0.121\pm0.13$ & $\underline{0.413\pm0.22}$ & $0.095\pm0.18$ \\
 & Overall & $0.547\pm0.35$ & $0.463\pm0.36$ & $0.146\pm0.23$ & $\underline{0.324\pm0.28}$ & $0.066\pm0.13$ \\
\addlinespace[1.5em]
\multirow{3}{*}{Homogeneous} & Attack & $0.567\pm0.42$ & $0.507\pm0.41$ & $\underline{\mathbf{0.053\pm0.14}}$ & $0.440\pm0.37$ & $\underline{\mathbf{0.000\pm0.00}}$ \\
 & Support & $\underline{\mathbf{0.534\pm0.30}}$ & $\underline{\mathbf{0.419\pm0.26}}$ & $\underline{\mathbf{0.111\pm0.18}}$ & $0.470\pm0.20$ & $\underline{\mathbf{0.000\pm0.00}}$ \\
 & Overall & $\underline{0.557\pm0.39}$ & $0.481\pm0.38$ & $\underline{\mathbf{0.070\pm0.15}}$ & $0.449\pm0.33$ & $\underline{\mathbf{0.000\pm0.00}}$ \\
\end{tabular}
\end{table*}

\vspace{3em}

\begin{table*}[htbp]
\centering
\caption{Retrieval performance on the \textbf{ArgTumour Arguments} dataset using \textbf{Qwen3-Embedding-8B}.}
\label{tab:results_argtumour_arguments_qwen3_embedding_8b}
\renewcommand{\arraystretch}{1.2}
\begin{tabular}{l c ccccc}
\toprule[1.5pt]
\textbf{Strategy} & \textbf{Stance} & \textbf{NDCG@10} & \textbf{Prec@R} & \textbf{Hard@R} & \textbf{Semi@R} & \textbf{Easy@R} \\
\midrule[1.5pt]
\multicolumn{7}{l}{\textbf{GROUP 1: Dense Retrieval}} \\
\multirow{3}{*}{Zero-Shot} & Attack & $0.705\pm0.38$ & $0.683\pm0.39$ & $0.183\pm0.29$ & $\underline{\mathbf{0.125\pm0.23}}$ & $0.009\pm0.04$ \\
 & Support & $0.782\pm0.19$ & $0.625\pm0.19$ & $0.218\pm0.19$ & $\underline{\mathbf{0.118\pm0.12}}$ & $0.038\pm0.13$ \\
 & Overall & $0.728\pm0.33$ & $0.666\pm0.34$ & $0.193\pm0.26$ & $\underline{\mathbf{0.123\pm0.20}}$ & $0.017\pm0.08$ \\
\addlinespace[1.5em]
\multirow{3}{*}{Homogeneous} & Attack & $0.771\pm0.23$ & $0.626\pm0.31$ & $0.050\pm0.18$ & $0.324\pm0.30$ & $\underline{\mathbf{0.000\pm0.00}}$ \\
 & Support & $0.783\pm0.15$ & $0.669\pm0.17$ & $0.006\pm0.03$ & $0.325\pm0.17$ & $\underline{\mathbf{0.000\pm0.00}}$ \\
 & Overall & $0.775\pm0.20$ & $0.639\pm0.28$ & $0.037\pm0.16$ & $0.324\pm0.27$ & $\underline{\mathbf{0.000\pm0.00}}$ \\
\end{tabular}
\end{table*}

\vspace{3em}

\begin{table*}[htbp]
\centering
\caption{Retrieval performance on the \textbf{TFU Evaluation Arguments} dataset using \textbf{BGE-Large}.}
\label{tab:results_tfu_evaluation_arguments_bge_large}
\renewcommand{\arraystretch}{1.2}
\begin{tabular}{l c ccccc}
\toprule[1.5pt]
\textbf{Strategy} & \textbf{Stance} & \textbf{NDCG@10} & \textbf{Prec@R} & \textbf{Hard@R} & \textbf{Semi@R} & \textbf{Easy@R} \\
\midrule[1.5pt]
\multicolumn{7}{l}{\textbf{GROUP 1: Dense Retrieval}} \\
\multirow{3}{*}{Zero-Shot} & Attack & $0.590\pm0.27$ & $0.411\pm0.30$ & $0.416\pm0.30$ & $\underline{0.107\pm0.19}$ & $0.066\pm0.16$ \\
 & Support & $0.766\pm0.26$ & $0.591\pm0.33$ & $0.296\pm0.31$ & $0.071\pm0.18$ & $0.042\pm0.13$ \\
 & Overall & $0.673\pm0.28$ & $0.497\pm0.33$ & $0.359\pm0.31$ & $0.090\pm0.19$ & $0.055\pm0.14$ \\
\addlinespace[1.5em]
\multirow{3}{*}{Homogeneous} & Attack & $0.734\pm0.29$ & $0.599\pm0.31$ & $\underline{\mathbf{0.104\pm0.20}}$ & $0.261\pm0.28$ & $0.036\pm0.12$ \\
 & Support & $0.775\pm0.31$ & $0.658\pm0.36$ & $\underline{\mathbf{0.147\pm0.28}}$ & $0.146\pm0.27$ & $0.049\pm0.15$ \\
 & Overall & $0.753\pm0.30$ & $0.627\pm0.33$ & $\underline{\mathbf{0.124\pm0.24}}$ & $0.207\pm0.28$ & $0.043\pm0.14$ \\
\end{tabular}
\end{table*}

\vspace{3em}

\begin{table*}[htbp]
\centering
\caption{Retrieval performance on the \textbf{TFU Evaluation Arguments} dataset using \textbf{Instructor-XL}.}
\label{tab:results_tfu_evaluation_arguments_instructor_xl}
\renewcommand{\arraystretch}{1.2}
\begin{tabular}{l c ccccc}
\toprule[1.5pt]
\textbf{Strategy} & \textbf{Stance} & \textbf{NDCG@10} & \textbf{Prec@R} & \textbf{Hard@R} & \textbf{Semi@R} & \textbf{Easy@R} \\
\midrule[1.5pt]
\multicolumn{7}{l}{\textbf{GROUP 1: Dense Retrieval}} \\
\multirow{3}{*}{Zero-Shot} & Attack & $0.740\pm0.22$ & $0.538\pm0.30$ & $0.413\pm0.29$ & $\underline{\mathbf{0.028\pm0.10}}$ & $0.021\pm0.09$ \\
 & Support & $0.757\pm0.27$ & $0.548\pm0.32$ & $0.408\pm0.32$ & $\underline{\mathbf{0.025\pm0.10}}$ & $0.018\pm0.08$ \\
 & Overall & $0.748\pm0.24$ & $0.543\pm0.31$ & $0.411\pm0.31$ & $\underline{\mathbf{0.027\pm0.10}}$ & $0.020\pm0.09$ \\
\addlinespace[1.5em]
\multirow{3}{*}{Homogeneous} & Attack & $\underline{\mathbf{0.797\pm0.25}}$ & $\underline{\mathbf{0.649\pm0.31}}$ & $\underline{\mathbf{0.196\pm0.27}}$ & $0.137\pm0.22$ & $0.018\pm0.08$ \\
 & Support & $\underline{\mathbf{0.861\pm0.27}}$ & $\underline{\mathbf{0.760\pm0.32}}$ & $\underline{\mathbf{0.152\pm0.29}}$ & $0.062\pm0.17$ & $0.026\pm0.10$ \\
 & Overall & $\underline{\mathbf{0.827\pm0.26}}$ & $\underline{\mathbf{0.702\pm0.32}}$ & $\underline{\mathbf{0.175\pm0.28}}$ & $0.101\pm0.21$ & $0.022\pm0.09$ \\
\end{tabular}
\end{table*}

\vspace{3em}

\begin{table*}[htbp]
\centering
\caption{Retrieval performance on the \textbf{TFU Evaluation Arguments} dataset using \textbf{Qwen3-Embedding-8B}.}
\label{tab:results_tfu_evaluation_arguments_qwen3_embedding_8b}
\renewcommand{\arraystretch}{1.2}
\begin{tabular}{l c ccccc}
\toprule[1.5pt]
\textbf{Strategy} & \textbf{Stance} & \textbf{NDCG@10} & \textbf{Prec@R} & \textbf{Hard@R} & \textbf{Semi@R} & \textbf{Easy@R} \\
\midrule[1.5pt]
\multicolumn{7}{l}{\textbf{GROUP 1: Dense Retrieval}} \\
\multirow{3}{*}{Zero-Shot} & Attack & $0.787\pm0.20$ & $0.581\pm0.30$ & $0.381\pm0.30$ & $\underline{\mathbf{0.025\pm0.09}}$ & $0.013\pm0.07$ \\
 & Support & $0.843\pm0.26$ & $0.680\pm0.33$ & $0.285\pm0.32$ & $\underline{\mathbf{0.021\pm0.10}}$ & $\underline{\mathbf{0.014\pm0.07}}$ \\
 & Overall & $0.814\pm0.23$ & $0.628\pm0.32$ & $0.335\pm0.32$ & $\underline{\mathbf{0.023\pm0.10}}$ & $\underline{\mathbf{0.014\pm0.07}}$ \\
\addlinespace[1.5em]
\multirow{3}{*}{Homogeneous} & Attack & $0.918\pm0.17$ & $0.835\pm0.24$ & $\underline{\mathbf{0.038\pm0.13}}$ & $0.114\pm0.21$ & $0.014\pm0.07$ \\
 & Support & $0.871\pm0.29$ & $0.813\pm0.32$ & $\underline{\mathbf{0.040\pm0.17}}$ & $0.121\pm0.27$ & $0.026\pm0.12$ \\
 & Overall & $0.896\pm0.23$ & $0.824\pm0.28$ & $\underline{\mathbf{0.039\pm0.15}}$ & $0.117\pm0.24$ & $0.020\pm0.09$ \\
\end{tabular}
\end{table*}

\vspace{3em}

\begin{table*}[htbp]
\centering
\caption{Retrieval performance on the \textbf{AVeriTeC Arguments} dataset using \textbf{BGE-Large}.}
\label{tab:results_averitec_arguments_bge_large}
\renewcommand{\arraystretch}{1.2}
\begin{tabular}{l c ccccc}
\toprule[1.5pt]
\textbf{Strategy} & \textbf{Stance} & \textbf{NDCG@10} & \textbf{Prec@R} & \textbf{Hard@R} & \textbf{Semi@R} & \textbf{Easy@R} \\
\midrule[1.5pt]
\multicolumn{7}{l}{\textbf{GROUP 1: Dense Retrieval}} \\
\multirow{3}{*}{Zero-Shot} & Attack & $0.457\pm0.30$ & $0.322\pm0.28$ & $0.315\pm0.29$ & $\underline{\mathbf{0.202\pm0.26}}$ & $0.161\pm0.24$ \\
 & Support & $0.617\pm0.33$ & $0.453\pm0.36$ & $0.300\pm0.36$ & $\underline{0.139\pm0.26}$ & $0.108\pm0.24$ \\
 & Overall & $0.530\pm0.33$ & $0.382\pm0.33$ & $0.308\pm0.33$ & $\underline{0.173\pm0.26}$ & $0.137\pm0.24$ \\
\addlinespace[1.5em]
\multirow{3}{*}{Homogeneous} & Attack & $0.590\pm0.31$ & $0.464\pm0.31$ & $\underline{\mathbf{0.059\pm0.15}}$ & $0.452\pm0.32$ & $0.024\pm0.09$ \\
 & Support & $0.636\pm0.38$ & $0.519\pm0.40$ & $\underline{\mathbf{0.065\pm0.20}}$ & $0.340\pm0.39$ & $0.075\pm0.22$ \\
 & Overall & $0.611\pm0.35$ & $0.489\pm0.35$ & $\underline{\mathbf{0.062\pm0.18}}$ & $0.401\pm0.36$ & $0.047\pm0.16$ \\
\end{tabular}
\end{table*}

\vspace{3em}

\begin{table*}[htbp]
\centering
\caption{Retrieval performance on the \textbf{AVeriTeC Arguments} dataset using \textbf{Instructor-XL}.}
\label{tab:results_averitec_arguments_instructor_xl}
\renewcommand{\arraystretch}{1.2}
\begin{tabular}{l c ccccc}
\toprule[1.5pt]
\textbf{Strategy} & \textbf{Stance} & \textbf{NDCG@10} & \textbf{Prec@R} & \textbf{Hard@R} & \textbf{Semi@R} & \textbf{Easy@R} \\
\midrule[1.5pt]
\multicolumn{7}{l}{\textbf{GROUP 1: Dense Retrieval}} \\
\multirow{3}{*}{Zero-Shot} & Attack & $0.625\pm0.27$ & $0.452\pm0.30$ & $0.346\pm0.30$ & $\underline{\mathbf{0.120\pm0.21}}$ & $0.081\pm0.18$ \\
 & Support & $0.689\pm0.31$ & $0.488\pm0.36$ & $0.383\pm0.36$ & $\underline{\mathbf{0.069\pm0.19}}$ & $0.060\pm0.19$ \\
 & Overall & $0.655\pm0.29$ & $0.469\pm0.33$ & $0.363\pm0.33$ & $\underline{\mathbf{0.097\pm0.20}}$ & $0.072\pm0.18$ \\
\addlinespace[1.5em]
\multirow{3}{*}{Homogeneous} & Attack & $0.637\pm0.31$ & $\underline{\mathbf{0.496\pm0.31}}$ & $\underline{\mathbf{0.122\pm0.22}}$ & $0.357\pm0.32$ & $\underline{\mathbf{0.025\pm0.10}}$ \\
 & Support & $0.737\pm0.35$ & $\underline{\mathbf{0.623\pm0.40}}$ & $\underline{\mathbf{0.079\pm0.23}}$ & $0.246\pm0.36$ & $0.052\pm0.18$ \\
 & Overall & $0.683\pm0.33$ & $\underline{\mathbf{0.554\pm0.36}}$ & $\underline{\mathbf{0.102\pm0.22}}$ & $0.306\pm0.34$ & $\underline{\mathbf{0.037\pm0.14}}$ \\
\end{tabular}
\end{table*}

\vspace{3em}

\begin{table*}[htbp]
\centering
\caption{Retrieval performance on the \textbf{AVeriTeC Arguments} dataset using \textbf{Qwen3-Embedding-8B}.}
\label{tab:results_averitec_arguments_qwen3_embedding_8b}
\renewcommand{\arraystretch}{1.2}
\begin{tabular}{l c ccccc}
\toprule[1.5pt]
\textbf{Strategy} & \textbf{Stance} & \textbf{NDCG@10} & \textbf{Prec@R} & \textbf{Hard@R} & \textbf{Semi@R} & \textbf{Easy@R} \\
\midrule[1.5pt]
\multicolumn{7}{l}{\textbf{GROUP 1: Dense Retrieval}} \\
\multirow{3}{*}{Zero-Shot} & Attack & $0.689\pm0.25$ & $0.497\pm0.30$ & $0.334\pm0.30$ & $\underline{\mathbf{0.116\pm0.21}}$ & $0.053\pm0.14$ \\
 & Support & $0.756\pm0.30$ & $0.571\pm0.37$ & $0.324\pm0.37$ & $\underline{\mathbf{0.065\pm0.19}}$ & $0.040\pm0.15$ \\
 & Overall & $0.720\pm0.28$ & $0.531\pm0.33$ & $0.330\pm0.33$ & $\underline{\mathbf{0.093\pm0.20}}$ & $0.047\pm0.15$ \\
\addlinespace[1.5em]
\multirow{3}{*}{Homogeneous} & Attack & $0.778\pm0.25$ & $0.642\pm0.30$ & $\underline{\mathbf{0.042\pm0.13}}$ & $0.295\pm0.29$ & $0.020\pm0.09$ \\
 & Support & $0.741\pm0.37$ & $0.658\pm0.40$ & $\underline{\mathbf{0.016\pm0.11}}$ & $0.289\pm0.39$ & $0.037\pm0.16$ \\
 & Overall & $0.761\pm0.31$ & $0.650\pm0.35$ & $\underline{\mathbf{0.030\pm0.12}}$ & $0.292\pm0.34$ & $0.028\pm0.13$ \\
\end{tabular}
\end{table*}

\clearpage
\section{Ablation Metrics Tables}
\label{app:ablation_metrics_tables}

The following tables present the word-ablation diagnostic metrics (RIS, RCS, and DI) across all evaluated datasets and models. Note that, similar to the retrieval results, the reported standard deviations ($\sigma$) reflect the variance across different evaluation queries and \textit{unseen} instructions.

\begin{table*}[htbp]
\centering
\caption{Ablation metrics on the \textbf{TFU Training Arguments (20\% held-out)} dataset. Results present the mean ($\mu$) and standard deviation ($\sigma$). Metrics include Relative Instruction Sensitivity (RIS), Relative Claim Sensitivity (RCS), and Directional Impact (DI) calculated for positive and hard negative pairs.}
\label{tab:ablation_general}
\renewcommand{\arraystretch}{1.2}
\begin{tabular}{ll cccc}
\toprule[1.5pt]
\textbf{Model} & \textbf{Strategy} & \textbf{RIS} & \textbf{RCS} & \textbf{DI (Pos)} & \textbf{DI (Hard Neg)} \\
\midrule[1.5pt]
\multirow{4}{*}{BGE-Large} & Zero-Shot & $0.195\pm0.10$ & $0.805\pm0.10$ & $-0.010\pm0.01$ & $-0.002\pm0.01$ \\
 & Homogeneous & $0.354\pm0.25$ & $0.646\pm0.25$ & $-0.123\pm0.16$ & $0.179\pm0.17$ \\
 & Mixed & $0.331\pm0.22$ & $0.669\pm0.22$ & $-0.092\pm0.13$ & $0.157\pm0.15$ \\
 & Mixed + Aug & $0.315\pm0.21$ & $0.685\pm0.21$ & $-0.077\pm0.12$ & $0.138\pm0.13$ \\
\addlinespace[1.5em]
\multirow{4}{*}{Instructor-XL} & Zero-Shot & $0.139\pm0.09$ & $0.861\pm0.09$ & $-0.004\pm0.01$ & $0.001\pm0.01$ \\
 & Homogeneous & $0.374\pm0.22$ & $0.626\pm0.22$ & $-0.085\pm0.10$ & $0.124\pm0.12$ \\
 & Mixed & $0.299\pm0.20$ & $0.701\pm0.20$ & $-0.040\pm0.06$ & $0.059\pm0.07$ \\
 & Mixed + Aug & $0.265\pm0.19$ & $0.735\pm0.19$ & $-0.031\pm0.04$ & $0.041\pm0.05$ \\
\addlinespace[1.5em]
\multirow{4}{*}{Qwen3-Embedding-8B} & Zero-Shot & $0.248\pm0.14$ & $0.752\pm0.14$ & $-0.008\pm0.03$ & $0.035\pm0.04$ \\
 & Homogeneous & $0.360\pm0.26$ & $0.640\pm0.26$ & $-0.199\pm0.21$ & $0.238\pm0.25$ \\
 & Mixed & $0.362\pm0.25$ & $0.638\pm0.25$ & $-0.164\pm0.18$ & $0.233\pm0.23$ \\
 & Mixed + Aug & $0.356\pm0.24$ & $0.644\pm0.24$ & $-0.147\pm0.16$ & $0.214\pm0.21$ \\
\bottomrule[1.5pt]
\end{tabular}
\end{table*}

\vspace{1em}

\begin{table*}[htbp]
\centering
\caption{Ablation metrics on the \textbf{TFU Validation Arguments generated by GPT-5} dataset. Results present the mean ($\mu$) and standard deviation ($\sigma$). Metrics include Relative Instruction Sensitivity (RIS), Relative Claim Sensitivity (RCS), and Directional Impact (DI) calculated for positive and hard negative pairs.}
\label{tab:ablation_validation}
\renewcommand{\arraystretch}{1.2}
\begin{tabular}{ll cccc}
\toprule[1.5pt]
\textbf{Model} & \textbf{Strategy} & \textbf{RIS} & \textbf{RCS} & \textbf{DI (Pos)} & \textbf{DI (Hard Neg)} \\
\midrule[1.5pt]
\multirow{4}{*}{BGE-Large} & Zero-Shot & $0.187\pm0.13$ & $0.813\pm0.13$ & $-0.009\pm0.01$ & $-0.003\pm0.01$ \\
 & Homogeneous & $0.362\pm0.25$ & $0.638\pm0.25$ & $-0.135\pm0.14$ & $0.171\pm0.16$ \\
 & Mixed & $0.318\pm0.21$ & $0.682\pm0.21$ & $-0.091\pm0.10$ & $0.137\pm0.12$ \\
 & Mixed + Aug & $0.295\pm0.20$ & $0.705\pm0.20$ & $-0.075\pm0.09$ & $0.117\pm0.11$ \\
\addlinespace[1.5em]
\multirow{4}{*}{Instructor-XL} & Zero-Shot & $0.134\pm0.09$ & $0.866\pm0.09$ & $-0.004\pm0.01$ & $0.001\pm0.01$ \\
 & Homogeneous & $0.389\pm0.21$ & $0.611\pm0.21$ & $-0.086\pm0.09$ & $0.125\pm0.10$ \\
 & Mixed & $0.320\pm0.20$ & $0.680\pm0.20$ & $-0.047\pm0.05$ & $0.064\pm0.07$ \\
 & Mixed + Aug & $0.284\pm0.19$ & $0.716\pm0.19$ & $-0.036\pm0.04$ & $0.047\pm0.05$ \\
\addlinespace[1.5em]
\multirow{4}{*}{Qwen3-Embedding-8B} & Zero-Shot & $0.287\pm0.13$ & $0.713\pm0.13$ & $0.006\pm0.03$ & $0.046\pm0.03$ \\
 & Homogeneous & $0.413\pm0.26$ & $0.587\pm0.26$ & $-0.205\pm0.19$ & $0.221\pm0.22$ \\
 & Mixed & $0.409\pm0.25$ & $0.591\pm0.25$ & $-0.161\pm0.15$ & $0.217\pm0.21$ \\
 & Mixed + Aug & $0.402\pm0.24$ & $0.598\pm0.24$ & $-0.148\pm0.14$ & $0.193\pm0.18$ \\
\bottomrule[1.5pt]
\end{tabular}
\end{table*}

\clearpage
\vspace{2em}

\begin{table*}[htbp]
\centering
\caption{Ablation metrics on the \textbf{TFU Validation Arguments generated by Qwen3-8B} dataset. Results present the mean ($\mu$) and standard deviation ($\sigma$). Metrics include Relative Instruction Sensitivity (RIS), Relative Claim Sensitivity (RCS), and Directional Impact (DI) calculated for positive and hard negative pairs.}
\label{tab:ablation_validationqwen}
\renewcommand{\arraystretch}{1.2}
\begin{tabular}{ll cccc}
\toprule[1.5pt]
\textbf{Model} & \textbf{Strategy} & \textbf{RIS} & \textbf{RCS} & \textbf{DI (Pos)} & \textbf{DI (Hard Neg)} \\
\midrule[1.5pt]
\multirow{4}{*}{BGE-Large} & Zero-Shot & $0.134\pm0.11$ & $0.866\pm0.11$ & $-0.007\pm0.01$ & $-0.002\pm0.01$ \\
 & Homogeneous & $0.323\pm0.24$ & $0.677\pm0.24$ & $-0.069\pm0.12$ & $0.157\pm0.17$ \\
 & Mixed & $0.283\pm0.21$ & $0.717\pm0.21$ & $-0.037\pm0.08$ & $0.130\pm0.14$ \\
 & Mixed + Aug & $0.251\pm0.20$ & $0.749\pm0.20$ & $-0.020\pm0.07$ & $0.106\pm0.11$ \\
\addlinespace[1.5em]
\multirow{4}{*}{Instructor-XL} & Zero-Shot & $0.103\pm0.07$ & $0.897\pm0.07$ & $-0.002\pm0.01$ & $0.003\pm0.01$ \\
 & Homogeneous & $0.341\pm0.20$ & $0.659\pm0.20$ & $-0.051\pm0.08$ & $0.115\pm0.10$ \\
 & Mixed & $0.273\pm0.19$ & $0.727\pm0.19$ & $-0.017\pm0.04$ & $0.067\pm0.07$ \\
 & Mixed + Aug & $0.238\pm0.18$ & $0.762\pm0.18$ & $-0.014\pm0.03$ & $0.049\pm0.06$ \\
\addlinespace[1.5em]
\multirow{4}{*}{Qwen3-Embedding-8B} & Zero-Shot & $0.280\pm0.13$ & $0.720\pm0.13$ & $-0.006\pm0.04$ & $0.051\pm0.04$ \\
 & Homogeneous & $0.400\pm0.26$ & $0.600\pm0.26$ & $-0.177\pm0.20$ & $0.214\pm0.23$ \\
 & Mixed & $0.400\pm0.25$ & $0.600\pm0.25$ & $-0.126\pm0.15$ & $0.218\pm0.22$ \\
 & Mixed + Aug & $0.384\pm0.23$ & $0.616\pm0.23$ & $-0.104\pm0.13$ & $0.195\pm0.19$ \\
\bottomrule[1.5pt]
\end{tabular}
\end{table*}

\vspace{1em}

\begin{table*}[htbp]
\centering
\caption{Ablation metrics on the \textbf{ArgTumour Arguments} dataset. Results present the mean ($\mu$) and standard deviation ($\sigma$). Metrics include Relative Instruction Sensitivity (RIS), Relative Claim Sensitivity (RCS), and Directional Impact (DI) calculated for positive and hard negative pairs.}
\label{tab:ablation_medical}
\renewcommand{\arraystretch}{1.2}
\begin{tabular}{ll cccc}
\toprule[1.5pt]
\textbf{Model} & \textbf{Strategy} & \textbf{RIS} & \textbf{RCS} & \textbf{DI (Pos)} & \textbf{DI (Hard Neg)} \\
\midrule[1.5pt]
\multirow{4}{*}{BGE-Large} & Zero-Shot & $0.152\pm0.08$ & $0.848\pm0.08$ & $-0.004\pm0.01$ & $-0.001\pm0.01$ \\
 & Homogeneous & $0.372\pm0.24$ & $0.628\pm0.24$ & $-0.090\pm0.13$ & $0.091\pm0.10$ \\
 & Mixed & $0.330\pm0.21$ & $0.670\pm0.21$ & $-0.036\pm0.10$ & $0.092\pm0.09$ \\
 & Mixed + Aug & $0.301\pm0.19$ & $0.699\pm0.19$ & $-0.030\pm0.09$ & $0.077\pm0.07$ \\
\addlinespace[1.5em]
\multirow{4}{*}{Instructor-XL} & Zero-Shot & $0.141\pm0.08$ & $0.859\pm0.08$ & $-0.004\pm0.01$ & $0.004\pm0.01$ \\
 & Homogeneous & $0.395\pm0.21$ & $0.605\pm0.21$ & $-0.076\pm0.12$ & $0.064\pm0.08$ \\
 & Mixed & $0.346\pm0.20$ & $0.654\pm0.20$ & $-0.039\pm0.08$ & $0.054\pm0.06$ \\
 & Mixed + Aug & $0.326\pm0.19$ & $0.674\pm0.19$ & $-0.043\pm0.08$ & $0.044\pm0.05$ \\
\addlinespace[1.5em]
\multirow{4}{*}{Qwen3-Embedding-8B} & Zero-Shot & $0.259\pm0.12$ & $0.741\pm0.12$ & $-0.002\pm0.03$ & $0.045\pm0.03$ \\
 & Homogeneous & $0.375\pm0.23$ & $0.625\pm0.23$ & $-0.221\pm0.17$ & $0.125\pm0.15$ \\
 & Mixed & $0.374\pm0.24$ & $0.626\pm0.24$ & $-0.174\pm0.13$ & $0.133\pm0.15$ \\
 & Mixed + Aug & $0.350\pm0.23$ & $0.650\pm0.23$ & $-0.143\pm0.11$ & $0.131\pm0.14$ \\
\bottomrule[1.5pt]
\end{tabular}
\end{table*}

\clearpage
\vspace{2em}

\begin{table*}[htbp]
\centering
\caption{Ablation metrics on the \textbf{TFU Evaluation Arguments} dataset. Results present the mean ($\mu$) and standard deviation ($\sigma$). Metrics include Relative Instruction Sensitivity (RIS), Relative Claim Sensitivity (RCS), and Directional Impact (DI) calculated for positive and hard negative pairs.}
\label{tab:ablation_evaluation}
\renewcommand{\arraystretch}{1.2}
\begin{tabular}{ll cccc}
\toprule[1.5pt]
\textbf{Model} & \textbf{Strategy} & \textbf{RIS} & \textbf{RCS} & \textbf{DI (Pos)} & \textbf{DI (Hard Neg)} \\
\midrule[1.5pt]
\multirow{4}{*}{BGE-Large} & Zero-Shot & $0.216\pm0.11$ & $0.784\pm0.11$ & $-0.004\pm0.02$ & $0.004\pm0.02$ \\
 & Homogeneous & $0.313\pm0.22$ & $0.687\pm0.22$ & $-0.053\pm0.14$ & $0.131\pm0.18$ \\
 & Mixed & $0.265\pm0.20$ & $0.735\pm0.20$ & $-0.022\pm0.10$ & $0.114\pm0.14$ \\
 & Mixed + Aug & $0.238\pm0.18$ & $0.762\pm0.18$ & $0.000\pm0.08$ & $0.098\pm0.12$ \\
\addlinespace[1.5em]
\multirow{4}{*}{Instructor-XL} & Zero-Shot & $0.133\pm0.08$ & $0.867\pm0.08$ & $0.000\pm0.01$ & $0.005\pm0.01$ \\
 & Homogeneous & $0.315\pm0.21$ & $0.685\pm0.21$ & $-0.022\pm0.08$ & $0.103\pm0.11$ \\
 & Mixed & $0.254\pm0.18$ & $0.746\pm0.18$ & $-0.004\pm0.04$ & $0.056\pm0.07$ \\
 & Mixed + Aug & $0.225\pm0.16$ & $0.775\pm0.16$ & $-0.008\pm0.03$ & $0.039\pm0.05$ \\
\addlinespace[1.5em]
\multirow{4}{*}{Qwen3-Embedding-8B} & Zero-Shot & $0.251\pm0.14$ & $0.749\pm0.14$ & $-0.006\pm0.04$ & $0.043\pm0.04$ \\
 & Homogeneous & $0.339\pm0.24$ & $0.661\pm0.24$ & $-0.162\pm0.19$ & $0.192\pm0.23$ \\
 & Mixed & $0.330\pm0.24$ & $0.670\pm0.24$ & $-0.125\pm0.16$ & $0.188\pm0.21$ \\
 & Mixed + Aug & $0.315\pm0.22$ & $0.685\pm0.22$ & $-0.104\pm0.14$ & $0.173\pm0.19$ \\
\bottomrule[1.5pt]
\end{tabular}
\end{table*}

\vspace{1em}

\begin{table*}[htbp]
\centering
\caption{Ablation metrics on the \textbf{AVeriTeC Arguments} dataset. Results present the mean ($\mu$) and standard deviation ($\sigma$). Metrics include Relative Instruction Sensitivity (RIS), Relative Claim Sensitivity (RCS), and Directional Impact (DI) calculated for positive and hard negative pairs.}
\label{tab:ablation_averitec}
\renewcommand{\arraystretch}{1.2}
\begin{tabular}{ll cccc}
\toprule[1.5pt]
\textbf{Model} & \textbf{Strategy} & \textbf{RIS} & \textbf{RCS} & \textbf{DI (Pos)} & \textbf{DI (Hard Neg)} \\
\midrule[1.5pt]
\multirow{4}{*}{BGE-Large} & Zero-Shot & $0.127\pm0.09$ & $0.873\pm0.09$ & $-0.006\pm0.01$ & $-0.002\pm0.01$ \\
 & Homogeneous & $0.311\pm0.21$ & $0.689\pm0.21$ & $-0.083\pm0.13$ & $0.103\pm0.17$ \\
 & Mixed & $0.275\pm0.19$ & $0.725\pm0.19$ & $-0.047\pm0.09$ & $0.098\pm0.14$ \\
 & Mixed + Aug & $0.228\pm0.17$ & $0.772\pm0.17$ & $-0.027\pm0.07$ & $0.074\pm0.10$ \\
\addlinespace[1.5em]
\multirow{4}{*}{Instructor-XL} & Zero-Shot & $0.107\pm0.07$ & $0.893\pm0.07$ & $-0.002\pm0.01$ & $0.002\pm0.01$ \\
 & Homogeneous & $0.312\pm0.19$ & $0.688\pm0.19$ & $-0.047\pm0.08$ & $0.093\pm0.11$ \\
 & Mixed & $0.231\pm0.16$ & $0.769\pm0.16$ & $-0.019\pm0.04$ & $0.046\pm0.06$ \\
 & Mixed + Aug & $0.197\pm0.14$ & $0.803\pm0.14$ & $-0.016\pm0.03$ & $0.030\pm0.05$ \\
\addlinespace[1.5em]
\multirow{4}{*}{Qwen3-Embedding-8B} & Zero-Shot & $0.217\pm0.12$ & $0.783\pm0.12$ & $-0.010\pm0.03$ & $0.024\pm0.03$ \\
 & Homogeneous & $0.363\pm0.23$ & $0.637\pm0.23$ & $-0.185\pm0.20$ & $0.159\pm0.23$ \\
 & Mixed & $0.354\pm0.23$ & $0.646\pm0.23$ & $-0.143\pm0.17$ & $0.161\pm0.22$ \\
 & Mixed + Aug & $0.339\pm0.21$ & $0.661\pm0.21$ & $-0.120\pm0.15$ & $0.144\pm0.19$ \\
\bottomrule[1.5pt]
\end{tabular}
\end{table*}

\clearpage
\section{Ablation Metrics Figures}
\label{app:ablation_metrics_figs}

\begin{figure}[!ht]
    \centering
    \includegraphics[width=1\linewidth]{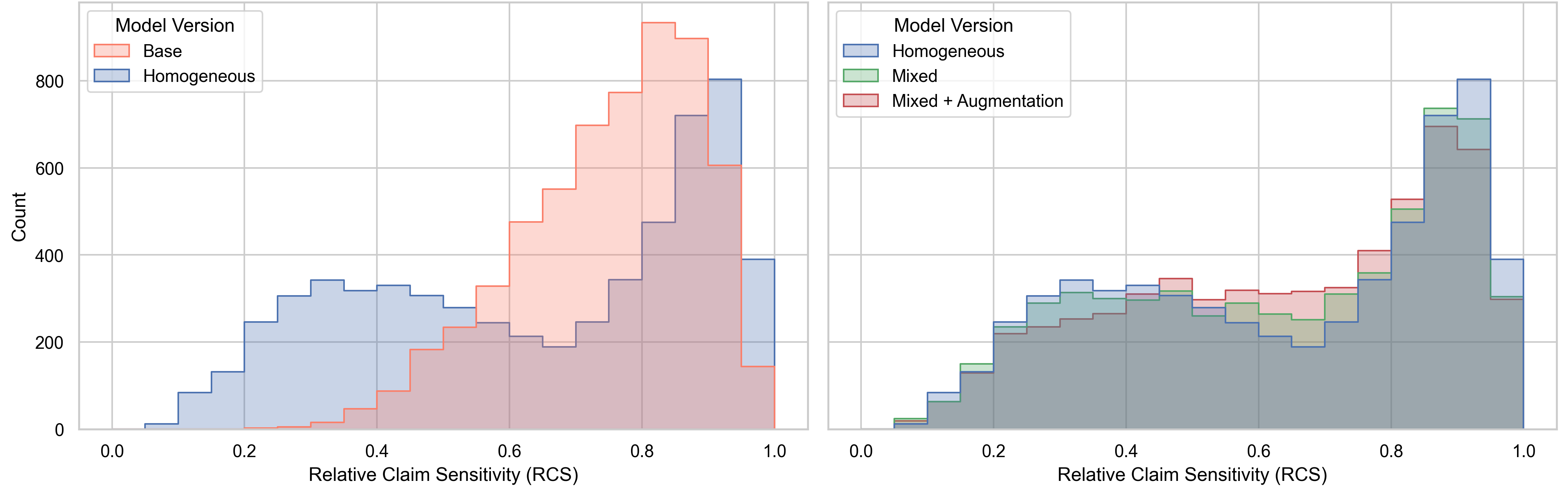}
    \caption{Relative Claim Sensitivity (RCS) distributions for Qwen3-Embedding-8B on the 20\% held-out TFU Training Arguments dataset. \textbf{Left}: The Base model exhibits a strong noun bias, focusing primarily on the claim (red distribution). The Homogeneous curriculum triggers topical collapse, shifting attention away from the claim to hyper-fixate on the instruction (blue distribution). \textbf{Right}: The data-centric interventions (Mixed and Mixed + Augmentation) successfully mitigate this collapse, flattening the secondary peak and organically restoring attention to the underlying topic.}
    \label{fig:train_rcs}
\end{figure}

\begin{figure}[!ht]
    \centering
    \includegraphics[width=1\linewidth]{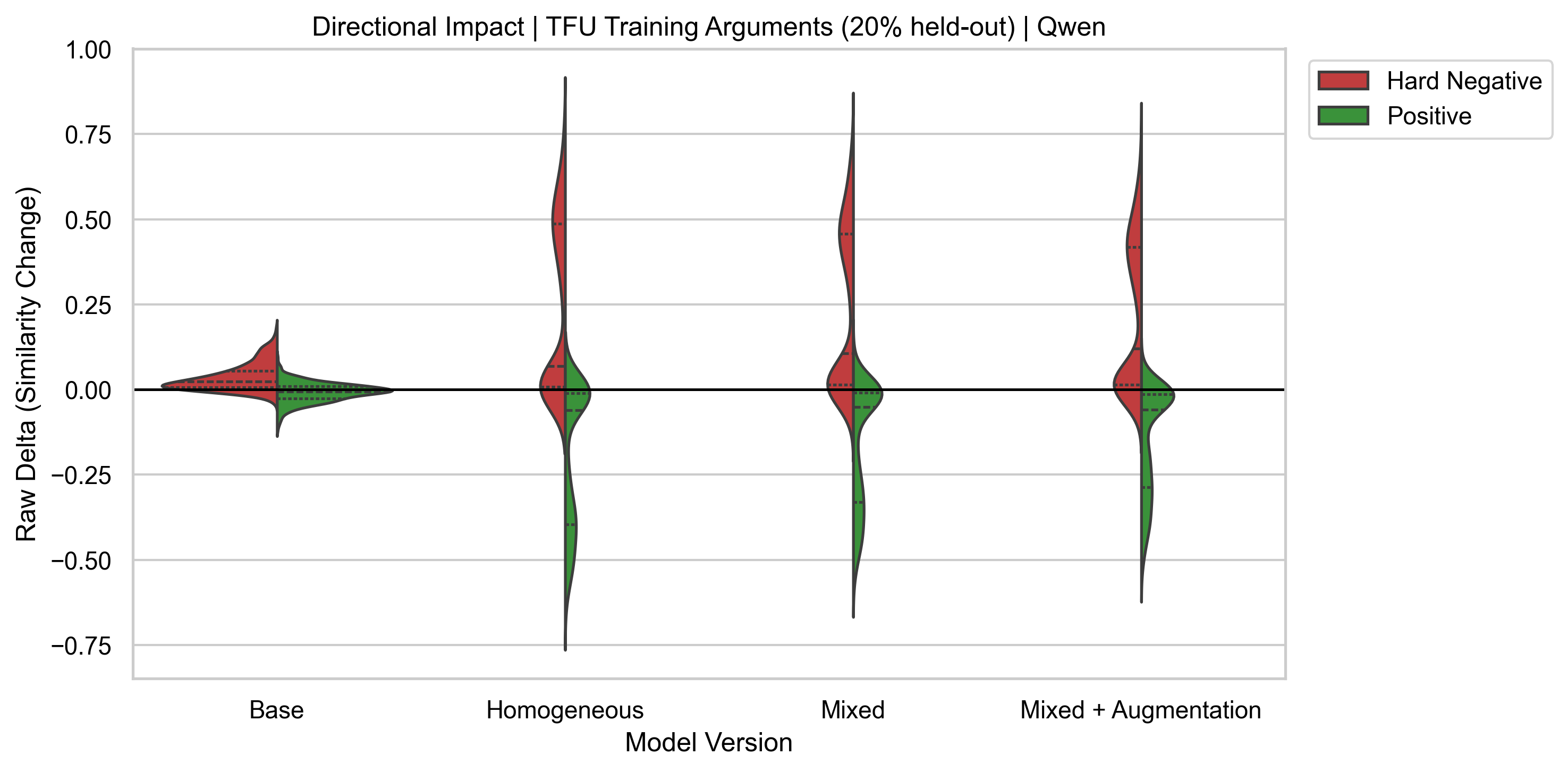}
    \caption{Directional Impact (DI) distributions for Qwen3-Embedding-8B evaluated on the 20\% held-out TFU Training Arguments dataset. The Base model exhibits near-zero similarity shifts when stance keywords are ablated, confirming its blindness to relational logic. Following fine-tuning, all training paradigms (Homogeneous, Mixed, and Mixed + Augmentation) demonstrate strong asymmetric stance boundaries, successfully pulling positive documents closer (green) while repelling stance-violating hard negatives (red).}
    \label{fig:train_di}
\end{figure}

\begin{figure}[!ht]
    \centering
    \includegraphics[width=1\linewidth]{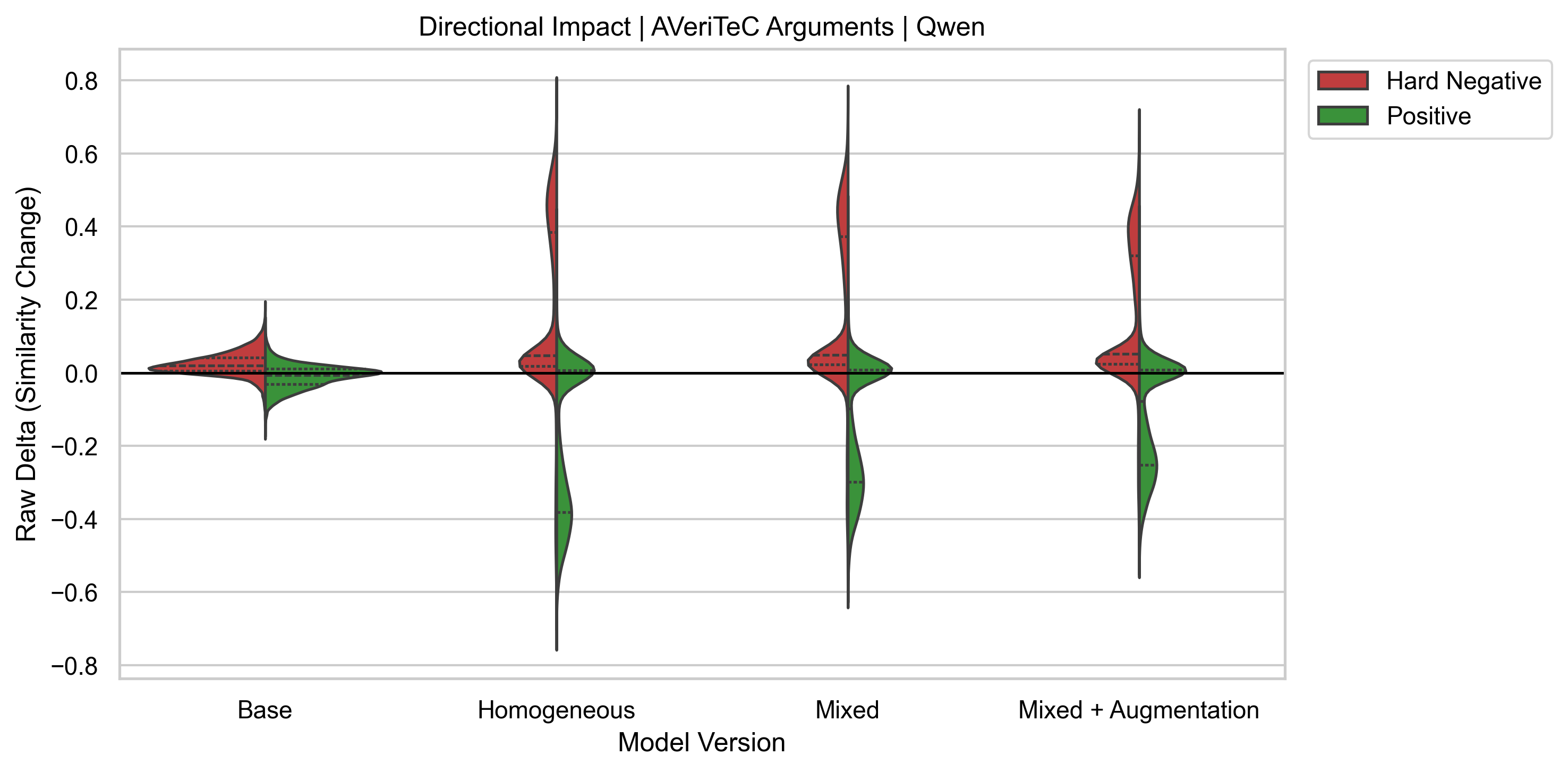}
    \caption{Directional Impact (DI) distributions for Qwen3-Embedding-8B on the out-of-domain AVeriTeC Arguments dataset. Mirroring the in-domain results, the Base model fails to process directional instructions. Crucially, the Mixed and Mixed + Augmentation models sustain their wide distributional splits between positive and hard-negative documents, showing that the data-centric interventions maintain strict, generalisable stance boundaries even when recovering from topical collapse.}
    \label{fig:averitec_di}
\end{figure}

\clearpage

\section{Word-Ablation Heatmap}
\label{app:ablation_heatmap}

\begin{figure}[!ht]
    \centering
    \includegraphics[width=1\linewidth]{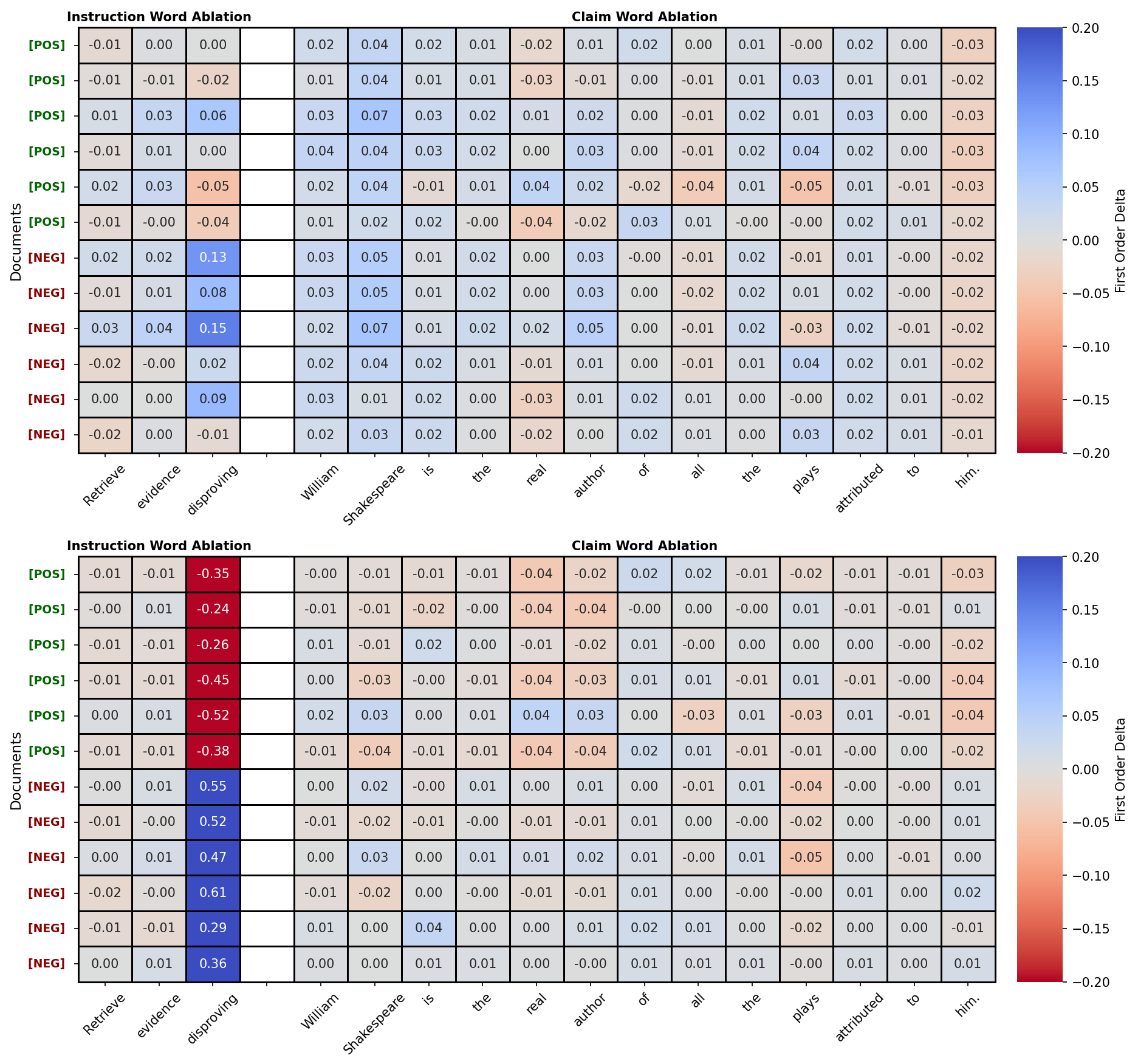}
    \caption{Word-ablation grids analysing the impact of removing individual words on document retrieval scores for base and tuned (homogeneous) Qwen3-Embedding-8B models. The $x$-axis displays the removed words from both the instruction and the claim. The $y$-axis represents the evaluation documents, separated into those attacking (\texttt{[POS]}) and supporting (\texttt{[NEG]}) the claim. The cell values and colours indicate the first-order delta in retrieval score when a specific word is ablated. The arguments (in-order) are given on the next page.}
    \label{fig:ablation_shakespeare}
\end{figure}

Word-level ablation heatmaps provide a direct, visual confirmation of the topical collapse phenomenon. Figure \ref{fig:ablation_shakespeare} isolates a specific query instance (``Retrieve evidence disproving'' + ``William Shakespeare is the real...'') on base and tuned (homogeneous) variants of Qwen3-Embedding-8B to illustrate how representational capacity is redistributed after fine-tuning. In the base model, the stance verb ``disproving'' exerts inconsistent and frequently minimal influence, confirming baseline instruction blindness.

Conversely, the tuned model causes a significant shift in the delta values. The instruction verb ``disproving'' transforms into a salient, high-contrast anchor, heavily penalising stance-violating documents while pulling correct ones closer. However, this hyper-fixation comes at the direct expense of the claim's semantic anchors. The influence of the topical nouns collapses to near-zero, visually corroborating the macro-level RIS/RCS shifts. Ultimately, the model successfully learns to enforce the ``disproving'' constraint but at the cost of reducing its ability to verify the underlying subject matter.

\newpage

\textbf{Instruction}: Retrieve evidence disproving

\textbf{Claim}: William Shakespeare is the real author of all the plays attributed to him.

\textbf{Positive (Attacking) Arguments}:
\begin{itemize}
    \item The claim is false because there is a major discrepancy between the known life of the Stratford man---a provincial businessman---and the vast, expert knowledge of law, foreign languages, court politics, and aristocratic pursuits detailed in the plays.
    \item The claim is false because for the most prolific writer of the age, there is a complete lack of a literary paper trail; no personal letters, manuscripts, or books from his library have ever been found, which is highly improbable.
    \item The claim is false because Shakespeare's detailed will makes no mention whatsoever of any literary assets, such as personal books, plays in progress, or manuscripts, an inexplicable omission for a professional writer.
    \item The claim is false because the six surviving signatures of Shakespeare are shaky and inconsistent, appearing more like the labored scrawl of a barely literate person than the hand of a literary master.
    \item The claim's assertion that he authored `all' the plays is false because modern textual analysis provides strong evidence that several plays, including `Titus Andronicus' and `Henry VIII,' were collaborations, meaning he was not the sole author.
    \item The claim is false because alternative candidates, such as Edward de Vere, Earl of Oxford, possessed the elite education, travel experience, and intimate knowledge of court life that are consistently and accurately reflected in the plays, making them a more plausible match for the author's profile.
\end{itemize}

\textbf{Hard Negative (Supporting) Arguments}:
\begin{itemize}
    \item The claim is true because Shakespeare's name was printed on the title pages of numerous plays published during his lifetime, establishing a direct, public link between the man and the works.
    \item The claim is true because Shakespeare's friends and fellow actors, Ben Jonson, John Heminge, and Henry Condell, explicitly identified him as the author in the First Folio, the definitive collection of his plays. They knew him personally and would have had no reason to perpetuate a fraud.
    \item The claim is true because financial and legal records confirm that William Shakespeare of Stratford was a shareholder and leading actor in the company that owned and performed the plays, directly connecting the man to the business of the works.
    \item The claim is true because no one during Shakespeare's life or for two centuries after his death expressed any doubt about his authorship. The idea of an alternative author is a modern theory unsupported by any contemporary evidence.
    \item The claim is true because Shakespeare's grammar school education in Stratford would have provided a rigorous foundation in Latin, rhetoric, and classical literature, which is sufficient to explain the knowledge in the plays without requiring a university degree or noble birth.
    \item The claim is true because arguments against his authorship are often rooted in classist snobbery, presuming that a commoner or a `glover's son' could not possess such genius, which is an assumption about creative ability, not a fact.
\end{itemize}

\clearpage

\section{Stance Inversion Prompt}
\label{app:stance_inversion_prompt}

Prompt submitted to Gemini 3.5 Flash to generate stance-inverted variants of existing arguments:\\[1.5mm]
\noindent
You are an expert NLP data engineer tasked with generating ``Synthetic Stance-Inversions'' for a contrastive learning dataset. Your goal is to take a human-written document that supports (or attacks) a claim and rewrite it to explicitly REVERSE its stance.\\[0.5mm]

\textbf{CRITICAL CONSTRAINTS:}
\begin{enumerate}
    \item \textbf{Minimum Edit Distance:} You must maintain a 90\%+ proportional lexical overlap with the original text. Keep all nouns, historical entities, subjects, and sentence structures completely identical.
    \item \textbf{The Relational Flip:} You must ONLY alter the relational verbs, polarity adjectives, or conjunctions to reverse the argumentative polarity.
    \item \textbf{Factual Integrity (No Hallucinations):} Do not invent false historical facts or corrupt entity knowledge (e.g., do not say a peace activist hated peace). Instead, sever the causal link (e.g., ``absence of evidence does not disprove...'').
    \item \textbf{No Explanatory Bloat:} You are strictly forbidden from inventing new reasons, rationalisations, or concluding clauses. Do not explain \textit{why} the claim is false/true. Simply negate the existing causal link.
    \item \textbf{Semantic Diversity (No Lazy Negation):} Do not rely exclusively on inserting ``not'' or ``does not'' before a verb, as this creates logical paradoxes and brittle syntax. Use semantic antonyms or direct verb inversions. Do not over-rely on ``Although'' templates.
    \item \textbf{Output Format:} You must output strictly valid JSON containing the new text.
\end{enumerate}

\textbf{EXAMPLE}
\begin{itemize}
    \item \textbf{Claim:} Atlantis was a real, technologically advanced civilization.
    \item \textbf{Original Document:} The claim is false because there is a complete absence of archaeological evidence---such as ruins, tools, pottery, or inscriptions---to support the existence of a large, technologically advanced civilization in 9,000 BCE.
    \item \textbf{Output:} 
    \begin{itemize}
        \item \textbf{attack\_flipped:} [``The claim is true because a complete absence of archaeological evidence---such as ruins, tools, pottery, or inscriptions---fails to definitively disprove the existence of a large, technologically advanced civilization in 9,000 BCE.'']
    \end{itemize}
\end{itemize}

\textbf{YOUR TASK}
You will be given a list of support and a list of attack for a given claim and should return the support\_flipped and the attack\_flipped lists.

\clearpage

\section{Hybrid Retrieval Results}
\label{app:hybrid_retrieval_results}

\begin{figure}[!ht]
    \centering
    \includegraphics[width=1\linewidth]{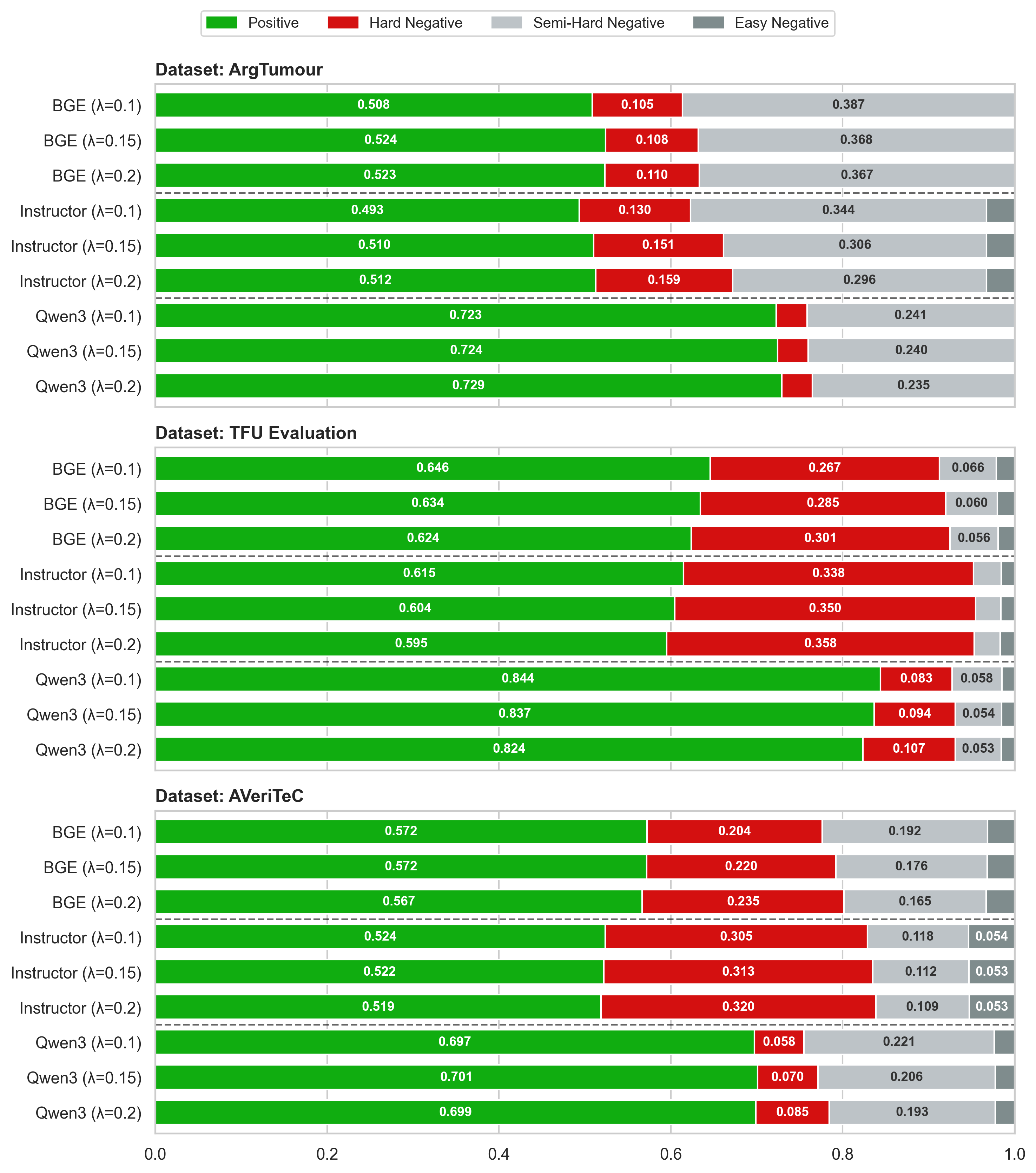}
    \caption{Hybrid retrieval performance combining the optimally tuned dense bi-encoder (Mixed + Augmentation) with a BM25 sparse retriever via Relative Score Fusion (RSF). The visualisations illustrate the impact of incrementally increasing the sparse weighting ($\lambda \in [0.1, 0.15, 0.2]$). Introducing the sparse retriever acts as a topical safety net, actively improving precision on highly specialised, entity-dense corpora like ArgTumour. However, because BM25 is fundamentally stance-blind, increasing its influence directly correlates with a higher number of stance errors across all architectures. This posits a strategic trade-off: hybrid search is highly effective for domain-specific entity matching but yields diminishing returns on general-domain datasets (e.g., TFU Evaluation) where the logic of the tuned dense model is already sufficient.}
    \label{fig:hybrid_search_results}
\end{figure}

\end{document}